\RequirePackage{fix-cm}

\documentclass[natbib,twocolumn]{svjour3}          
\smartqed  
\usepackage{graphicx}
\usepackage{xcolor}
\usepackage[pagebackref=true,breaklinks=true,citecolor=tealblue,colorlinks,linkcolor=ruby,bookmarks=false]{hyperref}
\definecolor{ruby}{rgb}{0.88, 0.07, 0.37}
\definecolor{tealblue}{rgb}{0.18, 0.40, 0.46}

\usepackage{array}
\usepackage{multirow}
\newcolumntype{H}{>{\setbox0=\hbox\bgroup}c<{\egroup}@{}}
\usepackage{amsmath}
\usepackage{amsfonts}

\usepackage{hyperref}
\usepackage{graphicx}
\usepackage{amssymb}
\usepackage{bm}
\usepackage{tabularx}
\usepackage{colortbl}

\usepackage{bbm}
\usepackage{dsfont}
\usepackage{pifont}

\usepackage{microtype}

\journalname{Noname}
\begin{document}

\title{CT3D++: Improving 3D Object Detection with Keypoint-induced Channel-wise Transformer}



\author{Hualian Sheng \and Sijia Cai \and Na Zhao \and Bing Deng \and Qiao Liang \and Min-Jian Zhao \and Jieping Ye
}


\institute{Hualian Sheng \and Sijia Cai  \and Bing Deng \and Qiao Liang \and Jieping Ye \at
              Alibaba Cloud, Hangzhou 310052, China \\
              \email{\{hualian.shl; stephen.csj;
dengbing.db;
chongchuan.lq; 
yejieping.ye\}@alibaba-inc.com}           
           \and
           Na Zhao \at
              Information Systems Technology and Design Pillar,
Singapore University of Technology and Design, 487372, Singapore\\
\email{na\_zhao@sutd.edu.sg}
\and
Min-Jian Zhao \at
College of Information Science and
Electronic Engineering, Zhejiang University, Hangzhou 310027, China\\
\email{mjzhao@zju.edu.cn}
}
\date{Received: date / Accepted: date}
\maketitle

\begin{abstract}
The field of 3D object detection from point clouds is rapidly advancing in computer vision, aiming to accurately and efficiently detect and localize objects in three-dimensional space. Current 3D detectors commonly fall short in terms of flexibility and scalability, with ample room for advancements in performance. In this paper, our objective is to address these limitations by introducing two frameworks for 3D object detection with minimal hand-crafted design. Firstly, we propose CT3D, which sequentially performs raw-point-based embedding, a standard Transformer encoder, and a channel-wise decoder for point features within each proposal. Secondly, we present an enhanced network called CT3D++, which incorporates geometric and semantic fusion-based embedding to extract more valuable and comprehensive proposal-aware information. Additionally, CT3D ++ utilizes a point-to-key bidirectional encoder for more efficient feature encoding with reduced computational cost. By replacing the corresponding components of CT3D with these novel modules, CT3D++ achieves state-of-the-art performance on both the KITTI dataset and the large-scale Waymo Open Dataset. The source code for our frameworks will be made accessible at \href{https://github.com/hlsheng1/CT3D-plusplus}{https://github.com/hlsheng1/CT3D-plusplus}.
\end{abstract}


\begin{figure}[t]
	\begin{center}
		\includegraphics[width=1.0\linewidth]{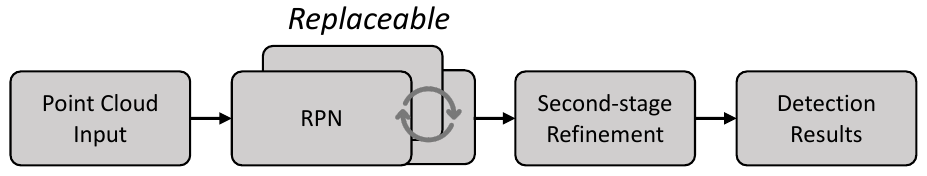}
	\end{center}
	\caption{General point cloud-based 3D object detection framework. We advocate to design flexible architecture with replaceable RPN.}
	\label{fig:ct3d_motivation}
\end{figure}

\section{Introduction}
3D object detection is a fundamental technology required for autonomous driving, which plays a vital role in driving decision-making and safety assurance. Many methods \cite{deng2021voxel,sheng2021improving,shi2020pv,yin2021center,li2021lidar,xu2022behind,he2022voxel} adopt a two-stage strategy that adds one extra refinement network based on the single-stage methods \cite{lang2019pointpillars,sheng2022rethinking,yan2018second,zheng2021cia,he2020structure} to achieve better performance. The main objective for the second-stage network is to effectively model the differences between the 3D proposals and ground-truths, and the main bottleneck is how to effectively extract the proposal-aware features.

Currently, refinement in two-stage methods \cite{shi2020pv,shi2023pv,deng2021voxel,li2021lidar} primarily relies on the PointNet \cite{qi2017pointnet} technique, which employs a flexible receptive field to aggregate features from the first-stage layers through a permutation-invariant network. However, these methods suffer from certain drawbacks due to their reliance on numerous hand-crafted designs, such as the neighbor-ball radii and grid size. Furthermore, these methods are tightly coupled with specific Region Proposal Network (RPN) architectures (such as SECOND \cite{yan2018second}), which include sparse convolution layers and conveniently provide mid-layer 3D voxel features \cite{shi2020pv,shi2023pv,deng2021voxel}. Without utilizing 3D voxel features, the performance of the refinement network is notably poor \cite{deng2021voxel,li2021lidar}. However, practical industry applications often require architectures that do not rely on 3D voxel features. Consequently, there is a need for a network structure that combines high performance with flexibility.

This paper aims to explore a more effective two-stage framework for 3D object detection that achieves high-quality refinement of 3D proposals while eliminating dependencies on specific RPN architectures. Taking inspiration from the Transformer network \cite{vaswani2017attention}, which excels at modeling unordered data formats, we propose a novel framework that enhances 3D object detection performance and enables the replacement of RPN with improved versions of the standard Transformer, as shown in Figure \ref{fig:ct3d_motivation}. Hence, we introduce a unified framework called CT3D/CT3D++, which leverages a novel channel-wise Transformer and a novel keypoint-indu\-ced channel-wise Transformer, respectively. Through our experiments, we demonstrate the effectiveness of our proposed methods in processing point cloud features, challenging previous conclusions regarding the necessity of using 3D voxel features for refinement based on PointNet \cite{qi2017pointnet}.

Firstly, we propose CT3D, an initial two-stage 3D object detection framework that only uses the raw point cloud in the second-stage network. This framework incorporates an arbitrary RPN to generate 3D proposals, followed by extracting per-proposal features from the raw point cloud. The second-stage network, referred to as the channel-wise Transformer, consists of three steps. Firstly, we propose a raw-point-based embedding to integrate proposal information into the raw points. Secondly, a standard Transformer encoder with a self-attention scheme is employed to capture point-wise interactions. Lastly, we introduce a novel channel-wise re-weighting approach to enhance the standard Transformer decoder for aggregating the encoded point features. This novel approach takes into consideration both global and local channel-wise features, resulting in improved expressiveness when assigning attention weights during query-key interactions. These three steps ensure the scalability of the 3D object detection framework while significantly enhancing detection performance.

Secondly, we propose an advanced 3D object detection framework called CT3D++. This framework is built upon CT3D and aims to generate more accurate detection results while reducing computational costs. Previous studies, such as PV-RCNN \cite{shi2020pv} and Voxel-RCNN \cite{deng2021voxel}, have highlighted that an excessive reliance on multi-scale 3D voxel features in the initial stage for gathering semantic information can impede model scalability and flexibility. Instead, an alternative and improved approach is to leverage other latent space features, such as Bird's Eye View (BEV) features. However, these investigations \cite{shi2020pv,deng2021voxel,yin2021center} have shown limited performance improvement when using BEV features. Therefore, we examine the combination of raw point geometric features and BEV semantic features to construct our refinement network, ensuring scalability, flexibility, and efficiency. Furthermore, in Section \ref{sec:abla}, we present ablation studies to demonstrate the redundancy of multi-scale 3D voxel features in our framework.

In the second stage of CT3D+++, we introduce the keypo\-int-induced channel-wise Transformer. This updated version incorporates enhancements to the embedding and encoder modules compared to its predecessor in CT3D. To expedite the process of raw point sampling, we propose a category-aware strategy that ensures a fixed capture area for points belonging to the same category. These sampled raw points are then projected onto the BEV feature map of the RPN, facilitating the extraction of abundant semantic information. In addition, we introduce a lightweight MLP network, comprising only a few hundred units, that operates on the concatenated geometric and semantic features. This integration significantly enhances the performance of object detection. Numerous existing studies have explored various networks, including PointNet \cite{li2021lidar}, Self-attention \cite{sheng2021improving}, and 
Convolutional Neural Network (CNN) \cite{shi2020pv,deng2021voxel}, to process fused features and generate the final output. However, these approaches lack efficient modeling of the relationship between the extracted features and the coarse 3D proposals, leading to diminished performance or high resource costs. Notably, the self-attenti\-on scheme exhibits quadratic computational complexity \cite{wang2020linformer}. To address this concern, we propose a more efficient and computationally economical alternative named as the point-to-key bidirectional cross-attention (PBC) scheme. In this scheme, we compute the correlation matrix between the raw points and the key points of the coarse 3D proposals, employing it as bidirectional attention weights to simultaneously update the raw and key point features. Our PBC scheme effectively models the interactions between raw points and key points while significantly reducing computational complexity compared to the self-attention scheme. Moreover, we validate in Section \ref{sec:abla} that our proposed scheme demonstrates superior performance.

This work makes several \textbf{key contributions}: (1). We introduce the CT3D framework, which includes two innovative modules: a raw-point-based embedding module for aggregating proposal information around each point, and a channel-wise re-weighting approach for more effective refinement of feature expression.
(2). We propose the CT3D++ framework, which incorporates two novel modules: a channel-wise attention-based fusion strategy for efficient combination of geometric and semantic features, and a point-to-key bidirectional cross-attention scheme that significantly reduces computational costs compared to the self-attention scheme while improving 3D bounding box refinement performance.
(3). Our CT3D/CT3D++ frameworks can be seamlessly integrated into existing RPNs, resulting in significant performance enhancements.
(4). Experimental results demonstrate that our CT3D/CT3D++ frameworks achieve state-of-the-art performance on both the Waymo Open Dataset and KITTI benchmark.

\section{Related Work}
\subsection{3D Object Detection from Image.}
Existing methods for 3D object detection from images can be broadly classified into two branches. The first branch focuses on monocular 3D object detection, where a single image is used to generate real-world bounding box information. The second branch is dedicated to multi-view 3D object detection, which relies on predictions from multiple images to estimate bounding boxes.

\subsubsection{Monocular 3D object detection} One approach in the field of monocular 3D object detection involves generating dense depth maps for image pixels, which are then used to estimate object-wise depth values. Several studies, such as AM3D \cite{ma2019accurate}, PatchNet \cite{ma2020rethinking}, Pseudo-LiDAR \cite{wang2019pseudo}, and PCT \cite{wang2021progressive}, have attempted to convert image pixels into pseudo point clouds using depth maps obtained from existing depth estimators \cite{cao2017estimating,song2021monocular,chen2021fixing}, combined with camera information. These pseudo point clouds are subsequently utilized by point cloud-based detectors to achieve 3D object detection. Specifically, the Pseudo-LiDAR approach \cite{wang2019pseudo} employs generated depth maps as a substitute for actual point clouds.
Alternatively, several other studies, including D$^4$LCN \cite{ding2020learning}, \cite{liu2021ground}, CaDDN \cite{reading2021categorical}, and MonoDETR \cite{zhang2023monodetr}, opt to combine depth maps with image features to produce the final 3D detection results. This approach relies on pixel-wise depth prediction, which has the potential to enhance 3D object detection. However, it may also pose challenges in terms of generalization.
Another group of studies bypasses the generation of pixel-wise depth information. Earlier works, such as SMOKE \cite{liu2020smoke} and FCOS3D \cite{wang2021fcos3d}, transform 2D object detectors into 3D object detectors by incorporating an additional depth estimation branch. However, their performance improvement is limited by the difficulties associated with long-range depth value predictions. Subsequently, numerous studies have focused on developing more efficient approaches to constrain the depth of object centers. Examples include MonoRCNN \cite{shi2021geometry}, Monoflex \cite{zhang2021objects}, GUPNet \cite{lu2021geometry}, Mo\-noPair \cite{chen2020monopair}, MonoCon \cite{liu2021learning}, and PDR \cite{sheng2023pdr}. These approaches leverage geometric information, such as prior knowledge or object relationships, resulting in significantly improved 3D object detection performance. Importantly, this group of methods typically requires fewer computational resources, making it the prevailing approach in current monocular 3D object detection technology.

\subsubsection{Multi-view 3D object detection} Multi-view 3D object detection has recently garnered significant attention for its ability to perform scene understanding using multiple surrounding cameras. Typically, this approach involves transforming image features from different camera views into explicit or implicit BEV feature maps. In explicit BEV-based methods, dense BEV feature maps are constructed using specific feature assignment strategies. Approaches like Lift-splat \cite{philion2020lift}, BEVDet \cite{huang2021bevdet}, and BEVDepth \cite{li2023bevdepth} spread image features onto the BEV map using predicted pixel-wise depth distributions. BEVFormer \cite{li2022bevformer} and BEVFormerV2 \cite{yang2023bevformer} aggregate image features through cross attention \cite{vaswani2017attention} between dense BEV features and image features. EVFormer \cite{li2022bevformer}, BEVFormerV2 \cite{yang2023bevformer}, and BEVDet4D \cite{huang2022bevdet4d} further incorporate temporal fusion by leveraging previous BEV frame information within the BEV feature map.
Implicit BEV-based methods, on the other hand, typically follow the strategy of predicting sparse location queries in BEV space, inspired by DETR \cite{carion2020end} technology. These methods then perform cross attention between these sparse BEV queries and image features to generate query-wise feature representations. Examples of such methods include DETR3D \cite{wang2022detr3d}, SparseBEV \cite{liu2023sparsebev}, and the PETR series \cite{liu2022petr,liu2023petrv2}. The implementation of cross attention in these studies may vary, primarily due to differences in feature sensing ranges and the utilization of distinct position embeddings.

\subsection{3D Object Detection from Point Cloud.}
In this section, we provide a brief review of closely related works in 3D detection based on LiDAR, categorized into single-stage and two-stage pipelines. Furthermore, we discuss works that leverage the Transformer architecture.

\subsubsection{Single-stage 3D object detection from point cloud}
Single-stage 3D detectors employ single-shot pipelines for efficient object detection. Typically, these methods perform 3D object detection on compressed Bird's Eye View (BEV) feature maps \cite{yan2018second,lang2019pointpillars,he2020structure,sheng2022rethinking,yang20203dssd}. Early approaches transform the 3D sparse point cloud data into compact 2D or 3D formats, such as range view-based methods \cite{chai2021point,fan2021rangedet,liang2021rangeioudet}, BEV view-based methods \cite{chen2017multi,ku2018joint}, and dense volume-based methods \cite{li20173d,yang2018pixor,zhou2018voxelnet}. Subsequently, traditional 2D/3D CNNs are employed to extract features. However, these methods often suffer from information loss or high computational costs. To address these challenges, SECOND \cite{yan2018second} introduces 3D sparse convolution to reduce the computational complexity associated with dense volume-based methods, while maintaining high efficiency. Building upon this, PointPillar \cite{lang2019pointpillars} replaces 3D sparse convolution with PointNet-like \cite{qi2017pointnet} feature extraction for each "pillar," resulting in faster detection speeds. SA-SSD \cite{he2020structure} introduces auxiliary branches for improved learning. CenterPoint \cite{yin2021center} incorporates an anchor-free detection head. CIA-SSD \cite{zheng2021cia} and AFDetV2 \cite{hu2022afdetv2} introduce IoU-guided confidence prediction. Additionally, RDIoU \cite{sheng2022rethinking} develops RDIoU-based optimization objectives to address the negative effects of rotation in 3D IoU-based optimization, leading to improved performance. Moreover, methods like 3DSSD \cite{yang20203dssd} directly perform predictions on raw point locations using PointNet \cite{qi2017pointnet} feature extraction.

\subsubsection{Two-stage 3D object detection from point cloud}
Two-stage 3D detectors enhance the performance and robustness of single-stage detectors by incorporating an additional refinement network. Improving the accuracy of 3D object detection is crucial as it directly impacts the security of industrial technical implementations. Generally, these methods can be classified into three categories: semantic information-based refinement (SIR), geometric information-based refinement (GIR), and hybrid information-based refinement (HIR). SIR methods leverage semantic information surrounding the  RPN's predictions. Examples include CenterPoint \cite{yin2021center} using BEV features and Voxel R-CNN \cite{deng2021voxel} and BtcDet \cite{xu2022behind} utilizing multi-level voxel features. These methods divide the 3D proposal into grids and extract semantic features from the RPN's layers based on the grid locations. GIR methods, on the other hand, focus on refining the 3D proposals using only raw point clouds. LiDAR-RCNN \cite{li2021lidar} introduces a PointNet-like network to process the points surrounding the proposals, while CT3D \cite{sheng2021improving} employs the Transformer architecture for refinement. These methods solely utilize geometric features around the proposals and achieve superior performance compared to those based on multi-level semantic features. Meanwhile, HIR methods combine both proposal-around raw points and multi-level semantic features. Examples include PointRCNN \cite{shi2019pointrcnn}, PV-RCNN \cite{shi2020pv}, and PV-RCNN++ \cite{shi2023pv}. These methods directly concatenate geometric and semantic features, but the performance improvement is limited. In contrast, we propose a novel framework that effectively leverages both SIR and GIR approaches to significantly enhance 3D object detection performance.

\subsubsection{3D object detection from point cloud using Transformer}
The Transformer \cite{vaswani2017attention} has garnered significant attention from researchers in the computer vision field, including applications in classification \cite{touvron2021training,dosovitskiy2020image} and 2D object detection \cite{carion2020end,liu2021swin}. In the domain of point cloud analysis, several studies have explored the use of Transformers for tasks such as classification, segmentation \cite{zhao2021point,guo2021pct}, and 3D object detection \cite{liu2021group,mao2021voxel,misra2021end,sheng2021improving,he2022voxel}. Group-free \cite{liu2021group} employs the Transformer to capture interactions among the predictions from the RPN. 3DETR \cite{misra2021end} treats point clouds as sequential data and produces 3D predictions in a manner similar to DETR \cite{carion2020end}. VoTR \cite{mao2021voxel} incorporates a self-attention scheme into the sparse convolution kernel. CT3D \cite{sheng2021improving} adopts an encoder-decoder Transformer architecture as the second-stage refinement module. VoxSet \cite{he2022voxel} models point cloud processing as a set-to-set translation using induced set Transformer \cite{lee2019set}, but its performance is limited by the induced latent codes. In this work, we propose an efficient Transformer that combines geometric and semantic features around the 3D proposals. Furthermore, we capitalize on the intrinsic keypoints of the 3D proposals to expedite the convergence rate and enhance the performance of the Transformer network.


\section{CT3D: Initial 3D Object Detection Framework with Channel-wise Transformer}

Most state-of-the-art methods for 3D object detection follow a two-stage framework that involves 3D region proposal generation and proposal feature refinement. It is worth noting that the popular RPN backbone \cite{yan2018second} achieves a recall rate of over 95\% on the KITTI 3D Detection Benchmark, while only achieving 78\% Average Precision (AP). This performance gap is attributed to the challenge of encoding objects and extracting robust features from 3D proposals, especially in cases of occlusion or long-range distances. Therefore, effectively modeling geometric relationships among points and leveraging accurate position information during the proposal feature refinement stage is critical for achieving good performance.
One important family of models is PointNet \cite{qi2017pointnet} and its variants \cite{qi2017pointnet++,pan20203d,shi2019pointrcnn}, which employ a flexible receptive field to aggregate features based on local regions and permutation-invariant networks. However, these methods suffer from the drawback of incorporating numerous hand-crafted designs, such as neighbor ball radii and grid sizes.
Another family of models is voxel-based methods \cite{yan2018second,song2016deep,zheng2021cia}, which utilize 3D convolutional kernels to extract information from neighboring voxels. However, the performance of such methods is suboptimal due to voxel quantization and sensitivity to hyperparameters. Subsequent studies \cite{zhou2018voxelnet,shi2020pv,deng2020voxel,he2020structure} have attempted to employ a point-voxel mixed strategy to capture multi-scale features while retaining fine-grained localization, but these approaches are closely tied to specific RPN architectures.

\begin{figure*}[t]
	\begin{center}
		\includegraphics[width=1.0\linewidth]{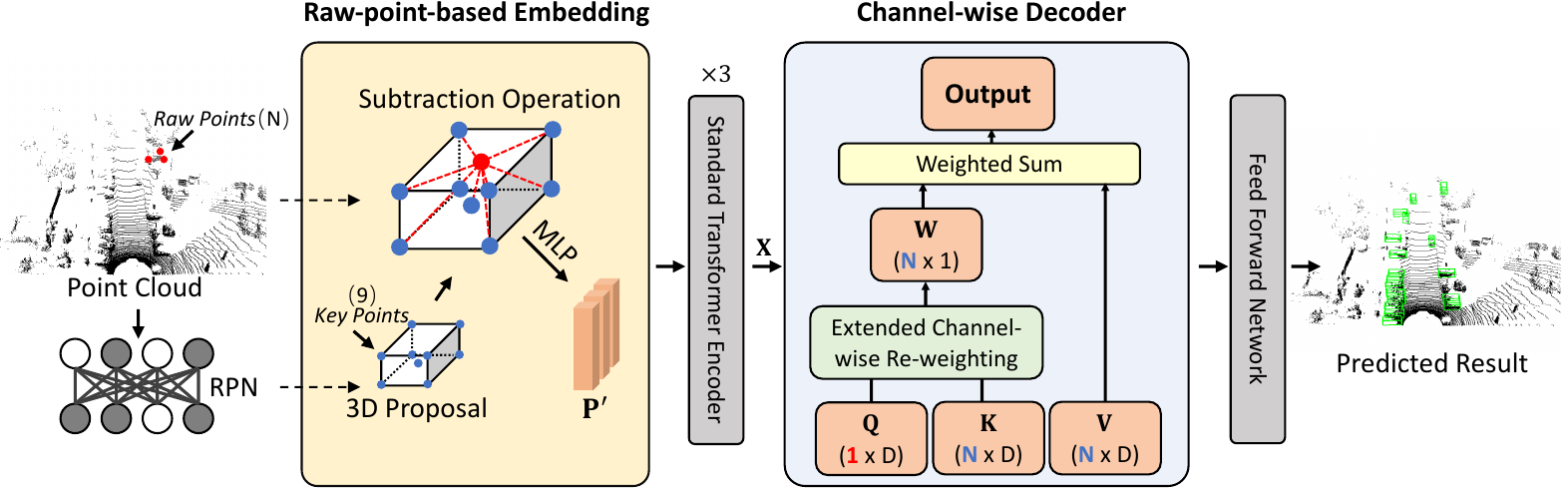}
	\end{center}
	\caption{\textbf{The overall framework of our proposed CT3D framework}. First, our CT3D utilizes an arbitrary RPN to generate coarse 3D proposals. Then, the raw points are gathered and processed using the proposed raw-point-based embedding module. Afterwards, the encoded point features  are transformed into an effective proposal feature representation by using three-layer standard Transformer encoder and one novel channel-wise decoder. Here, $\mathbf{K}$ and $\mathbf{V}$ are obtained by linear projection from $\mathbf{X}$. $\mathbf{Q}$ and $\mathbf{W}$ are learnable parameters.}
        \label{fig:ct3d_framework}
\end{figure*}

In this section, we provide a brief overview of our initial 3D object detection framework, CT3D \cite{sheng2021improving}, illustrated in Figure \ref{fig:ct3d_framework}. CT3D is a GIR method that solely utilizes raw points as the feature source for refining the first-stage proposals. Its second-stage network is a channel-wise Transformer, including a raw-point-based embedding, a standard Transformer encoder and a channel-wise decoder.


\subsection{Raw-point-based Embedding}\label{sec.ct3d_embedding}
The embedding layer is crucial within the Transformer architecture as it transforms input tokens into continuous dense vectors. For point tokens, it is essential to convert the absolute positional coordinates into standardized relative positional coordinates to ensure uniformity. Upon generating proposals via the RPN, we establish a scaled Region of Interest (RoI) within the point clouds corresponding to each proposal. This step aims to compensate for any deviation between the proposal and the corresponding ground-truth box by encompassing as many object points as possible. The scaled RoI area is specifically a cylindrical region with an infinite height and a radius $r = \alpha \sqrt{(\frac{l^c}{2})^2 + (\frac{w^c}{2})^2}$, where $\alpha$ serves as a hyper-parameter, and $l$ and $w$ represent the length and width of the proposal, respectively. Subsequently, we randomly sample $N = 256$ points from within the scaled RoIs, which are denoted as $\mathbf{P} = [\bm{p}_1, \dots, \bm{p}_N]$, for further processing.

Initially, we compute the relative coordinates between each sampled point and the centroid of the proposal to standardize the input distance feature, represented as $\Delta \bm{p}_i^c = \bm{p}_i - \bm{p}^c$. A straightforward approach is to directly concatenate the proposal information into each point feature, such as $[\Delta \bm{p}_i^c, l^c, w^c, h^c, \theta^c, f_i^r]$, where $f_i^r$ represents the raw point feature, such as reflection. However, employing a size-orien\-tation schema for representing the proposal yields only modest improvements in performance. This is attributed to the Transformer encoder's limited capability to reorient features effectively based on the specified geometric information.

It should be emphasized that keypoints frequently offer a more distinct geometric characteristic in detection tasks \cite{zhou2019bottom, law2018cornernet}. To capitalize on this attribute, we introduce an innovative technique termed \textbf{keypoints subtraction}, which calculates the relative coordinates between each point and the eight vertices of the associated proposal. These relative coordinates are calculated as $\Delta \bm{p}_i^j = \bm{p}_i - \bm{p}^j$, where $j = 1,\dots, 8$ and $\bm{p}^j$ represents the coordinates of the $j$-th corner point. 
Notably, the dimensions related to $l^c, w^c, h^c$, and $\theta^c$ are no longer present, as they are embedded within different components of the distance information. This approach allows the newly generated relative coordinates $\Delta \bm{p}_i^j$ to serve as a more informative representation of the proposal. After raw-point-based embedding, the proposal-guided point feature for each sampled point can be expressed as:
\begin{align}
	\bm{f}_i = \mathcal{A}([\Delta \bm{p}_i^c, \Delta \bm{p}_i^1, \dots, \Delta \bm{p}_i^8, f_i^r]) \in \mathbb{R}^{N\times D},
\end{align}
where $ \mathcal{A}(\cdot)$ is a linear projection layer. To simplify notation, we define $\mathbf{P}'=[\bm{f}_i], \forall i$ as the embedded point features.

\subsection{Standard Transformer Encoder}\label{sec.ct3d_encoder}
The embedded point features are then fed into the Transformer encoder with standard self-attention layers, followed by a feed-forward network (FFN) with residual structure. The self-attention mechanism enables neural networks to concentrate selectively on salient features within a given input context, enhancing the model's capability for point cloud scene interpretation. By weighting the significance of various input components, the self-attention mechanism facilitates a more nuanced understanding of the spatial relationships inherent in point cloud data. This, in turn, can lead to marked improvements in the network's performance on tasks requiring detailed contextual awareness. Subsequently, the embedded point features $\mathbf{P}'$ are encoded into $\mathbf{X}$.

\subsection{Channel-wise Decoder}\label{sec.ct3d_decoder}
We aim to decode all point features ($\mathbf{X}$) from the encoder module into a global representation, which is subsequently processed by FFNs for the final detection predictions. Unlike the standard Transformer decoder, which transforms multiple query embeddings using self-attention and encoder-dec\-oder attention mechanisms, our decoder only operates on a single query embedding based on the following observations:
(1). Processing a large number of proposals incurs high memory latency.
(2). Transforming multiple query embeddings independently would result in a corresponding number of words or objects, whereas our proposal refinement model requires a single prediction.

Generally, the final proposal representation obtained from the decoder can be seen as a weighted sum of all point features. Our primary objective is to determine dedicated decoding weights for each point. In the following sections, we begin by analyzing the standard cross-attention scheme and subsequently introduce an improved decoding scheme aimed at obtaining more effective decoding weights.

\begin{figure*}[t]
	\begin{center}
		\includegraphics[width=0.85\linewidth]{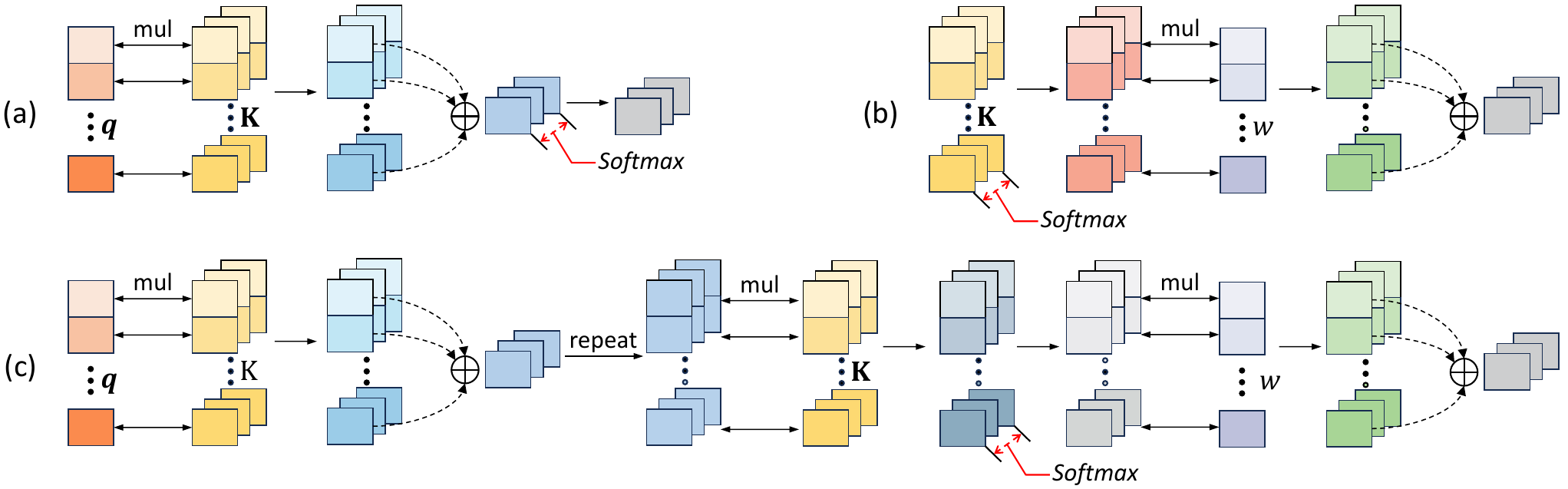}
	\end{center}
	\caption{Illustration of the different decoding schemes: (a) Standard decoding; (b) Channel-wise re-weighting; (c) Extended channel-wise re-weighting.}
	\label{fig:channel_wise}
\end{figure*}

\subsubsection{Standard Cross-attention} 
In the standard Transformer decoder, the conventional cross-attention mechanism employs a learnable vector, known as the query embedding, with a dimension of $D$ to amalgamate point features over all channels. Figure \ref{fig:channel_wise}(a) provides a graphical illustration of the resultant decoding weight vector for all point features within each attention head. For brevity, we exclude the head index when discussing multi-head attention, with the weight vector being calculated as described:
\begin{align}
    \bm{w}^{(S)} =\sigma\big(\frac{{\bm{q}}{\mathbf{K}}^T}{\sqrt{D}}\big),
\end{align}
where $\mathbf{K}$ represents the key embeddings obtained from the encoder output projection, while ${\bm{q}}$ is the corresponding query embedding. Each element of the product ${\bm{q}}{\mathbf{K}}^T$ can be interpreted as a global aggregation for each individual point (i.e., each key embedding). The \textit{softmax} function that follows assigns a decoding weight to each point based on the probability distribution within the normalized vector. However, this process results in decoding weights that are the outcome of a mere global aggregation and they do not account for local channel-specific modeling. Such local modeling is crucial for learning the 3D surface structures of point clouds, as different channels often have significant geometric interdependencies within point cloud data.

\subsubsection{Channel-wise Re-weighting} In order to emphasize the channel-wise information for key embeddings ${\mathbf{K}}^T$, a straightforward solution is to compute the decoding weight vector for points based on all the channels of ${\mathbf{K}}^T$. That is, we generate $D$ different decoding weight vectors for each channel to obtain $D$ decoding values. Further, a linear projection is introduced for these $D$ decoding values to form a united channel-wise decoding vector. As shown in Figure \ref{fig:channel_wise}(b), this new channel-wise re-weighting for decoding weight vector can be summarized as:
\begin{align}
    \bm{w}^{(C)} =\bm{s} \cdot {\sigma}\big(\frac{{\mathbf{K}}^T}{\sqrt{D}}\big), 
\end{align}
where $\bm{s}$ is a linear projection that compresses $D$ number of decoding values into a re-weighting scalar, ${\sigma}(\cdot)$ computes the \textit{softmax} along the $N$ dimension. However, the decoding weights computed by ${\sigma}(\cdot)$ are specific to each channel and do not consider the global aggregation of each point. Consequently, we can infer that the standard cross-attention mechanism in the Transformer decoder emphasizes global aggregation, while the channel-wise re-weighting approach focuses on local aggregation within each channel. To leverage the benefits of both approaches, we introduce an extended channel-wise re-weighting scheme as outlined below.

\begin{figure}[t]
	\begin{center}
		\includegraphics[width=1.0\linewidth]{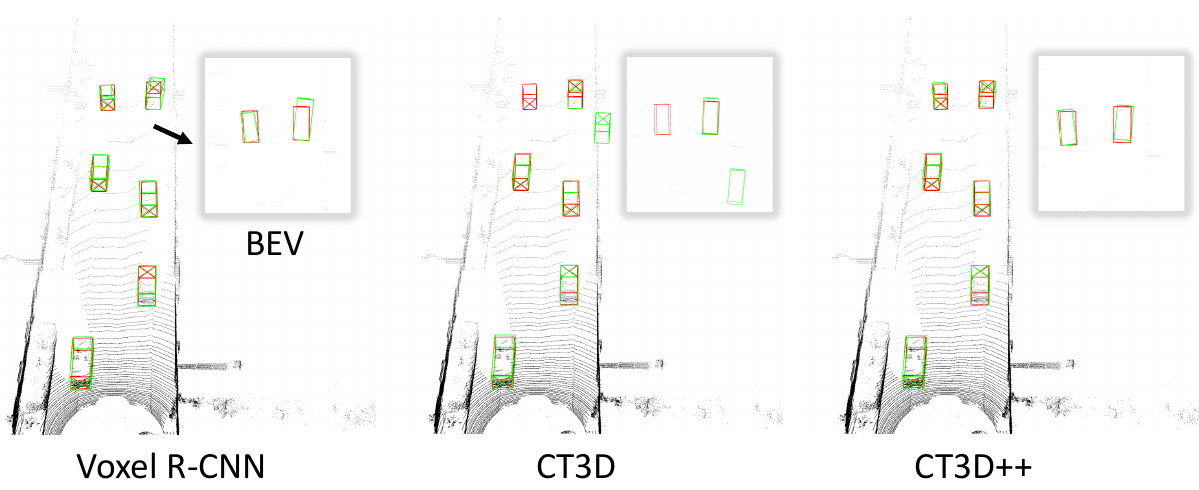}
	\end{center}
	\caption{The failure case analysis of Voxel R-CNN \cite{deng2021voxel} and our proposed CT3D. The predicted and ground-truth bounding boxes are shown in green and red, respectively. They generate biased bounding boxes and wrong confidence estimation, respectively. Instead, our newly proposed CT3D++ has good performance on these cases.}
	\label{fig:ct3d++_motivation}
\end{figure}

\subsubsection{Extended Channel-wise Re-weighting} 
To enhance the channel-wise re-weighting process, we propose a novel extended channel-wise re-weighting scheme. Specifically, we initiate the process by multiplying the query embedding with the key embeddings, allowing the spatial information to be distributed and replicated across all channels.
Subsequently, the resulting output is element-wise multiplied with the key embeddings, ensuring the preservation of channel distinctions. This extended channel-wise re-wei\-ghting scheme, illustrated in Figure \ref{fig:channel_wise}(c), generates the following decoding weight vector for all points:
\begin{align}
	\bm{w}^{(EC)} = \bm{s}\cdot{\sigma}\big(\frac{\rho({\bm{q}}{\mathbf{K}}^T)\odot {\mathbf{K}}^T}{\sqrt{D}}\big), 
\end{align}
where $\rho(\cdot)$ is a repeat operator makes $\mathbb{R}^{1\times N} \to \mathbb{R}^{D\times N}$. Through this approach, we preserve the global information akin to the channel-wise re-weighting scheme while simultaneously enhancing local and detailed channel interactions beyond what is possible with the standard decoding scheme. Moreover, this extended channel-wise re-weighting only br\-ings 1K+ (Bytes) increase as compared to the other two schemes. As a result, the final decoded proposal representation can be described as follows:
\begin{align}
    \bm{y} = \bm{w}^{(EC)}\cdot{\mathbf{V}},
\end{align}
where the value embeddings ${\mathbf{V}}$ is the linear projection obtained from ${\mathbf{X}}$.

\subsection{Training Losses}\label{sec.ct3d_losses}
The entire training loss consists of one first-stage RPN loss and one second-stage refinement loss. In the second stage, 
the detect head loss is a summation of a confidence loss and a regression loss. Following PV-RCNN \cite{shi2020pv}, we sample $\hat{M}$ anchors with 1:1 foreground and background ratio, and the confidence targets $\{\hat{u}_i\}_{i=1}^{\hat{M}}$
are the scaled values based on the 3D IoU values $\{o_i\}_{i=1}^{\hat{M}}$ between the 3D proposals and the ground truths. The whole detect head loss can be expressed as:
\begin{equation}
    \mathcal{L}_{Re}=\frac{1}{\hat{M}}[\sum_{i=1}^{\hat{M}}\mathcal{L}_{b}(\hat{a}_i, \hat{u}_i) + \mathds{1}(\hat{z})\mathcal{L}_{r}(\hat{b}_i), \hat{v}_i],
\end{equation}
where $\hat{z}$ represents the condition of $o_i > 0.55$, and $\mathcal{L}_{b}$ is implemented by binary cross entropy loss.
Finally, CT3D is trained by the combination of RPN loss $\mathcal{L}_{RPN}$ and second-stage refinement loss $\mathcal{L}_{Re}$.

\begin{figure*}[t]
	\begin{center}
		\includegraphics[width=1.0\linewidth]{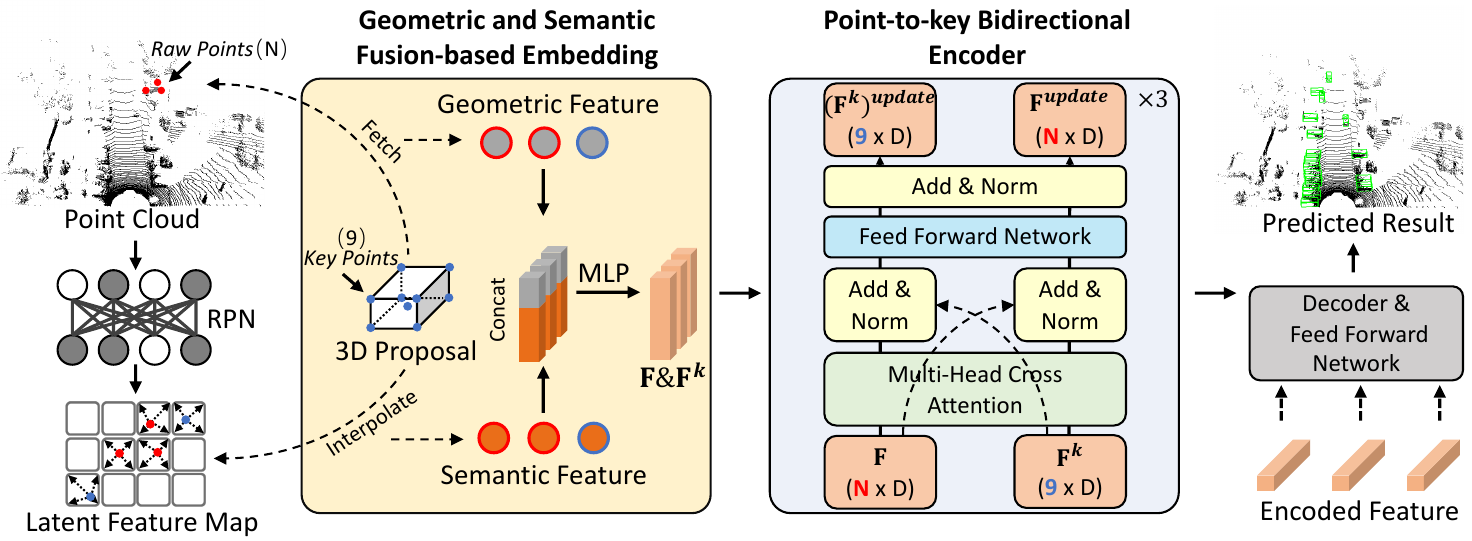}
	\end{center}
	\caption{\textbf{The overall framework of our proposed CT3D++ framework}. First, our CT3D++ utilizes an arbitrary RPN to generate coarse 3D proposals and a latent feature map. Then, the raw points and BEV features are gathered based on the 3D proposals using the proposed geometric and semantic fusion embedding module. After that, an efficient and low-cost point-to-key bidirectional cross-attention scheme is proposed to improve the point features and assign more attention to the foreground points. Finally, the encoded point features are decoded and used to predict the refined 3D object detection results.}
        \label{fig:ct3d++_framework}
\end{figure*}

In summary, CT3D represents one of the pioneering works that employ Transformer technology for point cloud-based 3D object detection. It offers improved architectural flexibility and reduced computational costs compared to previous methods in this domain. Nevertheless, there are areas within the CT3D architecture that can be further improved, specifically in terms of second-stage input feature selection and optimizing the high-cost self-attention mechanism.

\section{CT3D++: Faster and More Efficient Channel-wise Transformer for 3D Object Detection}
Our CT3D method attains cutting-edge performance, surpassing previous methodologies \cite{sheng2021improving}; however, it encounters limitations in effectively processing large-scale point clouds. This limitation originates from CT3D's reliance on a GIR framework in its second stage, which can often result in a deficiency of semantic information and consequently, an escalation in false positive predictions.
In Figure \ref{fig:ct3d++_motivation}, we compare CT3D with another existing method, Voxel-RCNN, which is an SIR method. The failure cases in Voxel-RCNN often involve biased bounding boxes, whereas the failure cases in CT3D frequently result in incorrect confidence estimations. Although some existing HIR methods \cite{shi2019pointrcnn,shi2020pv,shi2023pv} aim to combine the strengths of both SIR and GIR methods, their performance is often limited by the grid set abstraction operation used for downsampling. Additionally, their RPNs are fixed due to multi-scale voxel feature extraction, which restricts their applicability to real autonomous driving scenarios that typically rely on RPNs without voxel features, such as PointPillar \cite{lang2019pointpillars}. To the best of our knowledge, there is currently no efficient HIR method that fully exploits the advantages of both SIR and GIR while avoiding these limitations.

To explore a more efficient HIR method, we aim to enhance CT3D by improving its channel-wise Transformer component. The challenge lies in maintaining the flexibility of the overall architecture while effectively incorporating latent space features. It is worth noting that many single-stage approaches for point clouds, such as SECOND \cite{yan2018second}, PointPillar \cite{lang2019pointpillars}, SA-SSD \cite{he2020structure}, CIA-SSD \cite{zheng2021cia}, and RDIoU \cite{sheng2022rethinking}, typically compress point cloud features into a Bird's Eye View (BEV)-dimensional feature map. This raises the question of \textbf{whether we can extend CT3D by incorporating the compressed BEV features into the channel-wise Transformer input}.


Therefore, to ensure scalability, flexibility, and high performance, we investigate the combination of raw point geometric features and BEV semantic features for constructing our refinement network. In Section \ref{sec:experiments}, we present ablation studies to demonstrate the redundancy of multi-scale 3D voxel features in our framework. Specifically, we propose a category-aware raw point sampling strategy to accelerate the sampling process by fixing the capture area for points of the same category. These sampled raw points are then projected into the BEV feature map of the RPN to extract rich semantic information. Moreover, we introduce a lightweight MLP network with only a few hundred units. This MLP network is placed atop the concatenated geometric and semantic features, resulting in a significant enhancement in detection performance. The entire framework of our proposed CT3D++ is visually depicted in Figure \ref{fig:ct3d++_framework}. In comparison to CT3D, CT3D++ introduces two new modules that can replace their counterparts in CT3D, formed keypoint-induced channel-wise Transformer in the second stage. The first module is the geometric and semantic fusion-based embedding, which efficiently combines geometric and semantic point cloud features as the input for the second-stage Transformer. The second module is the bidirectional point-to-key encoder, which enables more efficient and effective context feature interaction.


\subsection{Geometric and Semantic Fusion-based Embedding}\label{sec.ct3d++_embedding}
We extract the geometric and semantic information from the raw point cloud and BEV feature map, respectively. These two sets of information are then fused and used as the input embedding for the Transformer encoder.

\subsubsection{Category-based raw point sampling} We sample the raw points around per bounding box in a cylindrical area with fixed radii according to different categories. This is because the size of the same category in the 3D real-world scene do not vary largely. Note that the point sampling with fixed radii can be accelerated by CUDA parallel processes. Hereinafter, we get $N$ sampled points $\mathbf{P}\in \mathbb{R}^{N\times 3}$ that are directly collected by LiDAR sensor with precise geometric information.

\subsubsection{Point-based BEV feature extraction} We project the sampled points $\mathbf{P}$ into the  down-sampled BEV feature map, and utilize the bilinear interpolation to get the interpolated features for each point, denoted as $\mathbf{T}$. The BEV feature map is the last layer of RPN architecture, and thus contains rich semantic information.

\subsubsection{Geometric and semantic fusion}
For each point, we fuse its geometric and semantic information by applying a simple MLP network $\mathcal{M}(\cdot)$, this process can be expressed as:
\begin{align}
    \mathbf{F} = \mathcal{M}(\mathcal{S}(\mathbf{P}), \mathbf{T}) \in \mathbb{R}^{N\times D},
\end{align}
where $\mathcal{S}(\cdot)$ is an inter-point embedding operation \cite{sheng2021improving} that normalizes the point coordinates via 3D coarse bounding boxes and $\mathcal{M}(\cdot)$ projects its input feature into dimension of $D$. Similarly, we get the fused features of nine key points of per coarse 3D bounding box, and denote it as $\mathbf{F}^k \in \mathbb{R}^{9\times D}$.

\subsection{Point-to-key Bidirectional Encoder}\label{sec.ct3d++_encoder}
The goal of the attention scheme is to improve the feature representation and assign more attention weights to the foreground points. Its core is to compute the correlation matrix among the inputs. We introduce the \textbf{Point-to-Key Bidirectional Cross-Attention (PBC)} as an attention mechanism that outperforms the conventional self-attention scheme \cite{vaswani2017attention}. Prior to delving into the intricacies of PBC, we shall elucidate the fundamentals of the self-attention mechanism for context.

\begin{figure}[t]
	\begin{center}
		\includegraphics[width=0.9\linewidth]{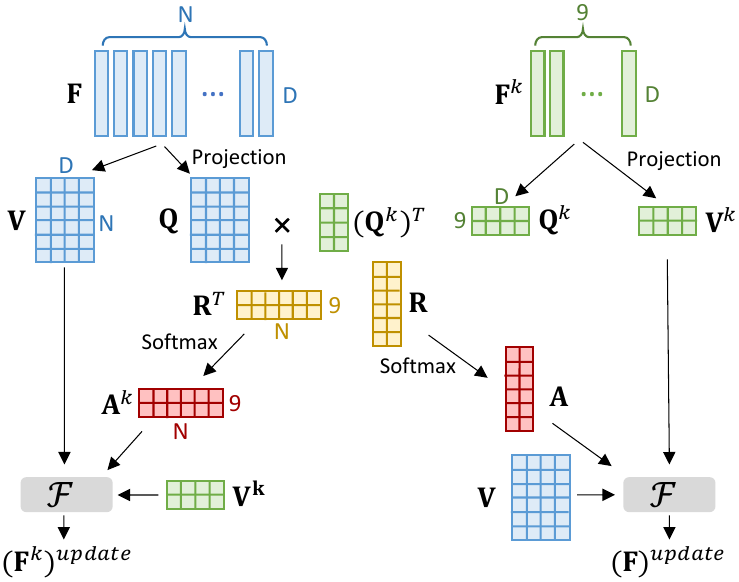}
	\end{center}
	\caption{Illustration of our proposed point-to-key bidirectional cross-attention scheme.}
	\label{fig:ct3d++_encoder}
\end{figure}

\subsubsection{Self-attention} 
It is designed to assign different attention weights to the input sequence, such that to enhance the feature representation of key information while suppress the interference of non-key information. It converts the input into three representations $\mathbf{Q}, \mathbf{K}, \mathbf{V}$ via learnable linear projection, and then computes the attention matrix:
\begin{align}
    \mathbf{A} = \sigma(\frac{\mathbf{Q}\mathbf{K}^T}{\sqrt{D}}),
\end{align}
where $\sigma(\cdot)$ computes the \textit{softmax} along the dimension N. After that, it multiplies attention matrix $\mathbf{A}$ and projected input matrix $\mathbf{V}$ to update the input features:
\begin{align}
    \mathbf{F}^{update} = \mathcal{F}(\mathbf{A}\cdot \mathbf{V} + \mathbf{V}),
\end{align}
where $\mathcal{F}(\cdot)$ is a simple feed forward network with residual connection. 

\noindent\textit{Analysis.} Enhancement of extracted point cloud features with a self-attention mechanism can substantially improve the representation's expressiveness and facilitate the modeling of inter-point relational features. However, this approach to feature processing indiscriminately addresses all input points, presenting formidable challenges in terms of convergence when learning the ground truth, as well as in reconciling discrepancies between the ground truth and the 3D proposals. Such convergence issues are especially pronounced in large-scale datasets characterized by substantial variability in the scale and positional information of foreground objects, potentially leading to a marked decline in performance.

\subsubsection{Our PBC} Towards the aforementioned convergence issue of the standard self-attention scheme, we propose PBC to enable more simplified and efficient attention-based feature updation. A few methods, \textit{e.g.,} Linformer \cite{wang2020linformer} and clustered attention \cite{vyas2020fast} propose to 
project the input features into the several clustering centers, however, their performance is limited by the unstable learning process of these clustering centers. In contrast, key points of 3D proposals in our method are inherent clustering points that can be used for complexity reduction.
Consequently, we propose \textbf{\textit{point-to-key bidirectional cross-attention}} to replace the self-attention in Transformer encoder  as shown in Figure \ref{fig:ct3d++_encoder}. 
Given the fused features $\mathbf{F}$ and $\mathbf{F}^k$, we respectively convert them into two separate linear projected features $\mathbf{Q}$, $\mathbf{V}$  and $\mathbf{Q}^k$, $\mathbf{V}^k$. We first compute the correlation matrix based on $\mathbf{Q}$ and $\mathbf{Q}^k$ as:
\begin{align}
    \mathbf{R} = \mathbf{Q}\cdot (\mathbf{Q}^k)^T.
\end{align}
Then, we can obtain the bidirectional attention matrices that respectively represent the assigned attention values for raw points and key points by computing:
\begin{align}
    \mathbf{A} = \sigma(\frac{\mathbf{R}}{\sqrt{D}}),\ \ \ \mathbf{A}^k = \hat{\sigma}(\frac{\mathbf{R}^T}{\sqrt{D}}),
\end{align}
where $\hat{\sigma}$ computes the \textit{softmax} along the dimension $9$. Both the sampled point features and the key point features are updated by their corresponding bidirectional attention matrices. 
After applying the attention-based aggregation for the input features, we adopt a simple FFN $\mathcal{F}(\cdot)$ with residual connection, yielding:
\begin{align}
    \mathbf{F}^{update} &= \mathcal{F}(\mathbf{A}\cdot \mathbf{V}^k + \mathbf{V}),\\
    (\mathbf{F}^{k})^{update} &= \mathcal{F}(\mathbf{A}^k\cdot \mathbf{V} + \mathbf{V}^k).
\end{align}
The above encoder layer can be repeated and stacked into a deep encoder module. We empirically find that three encoder layers give a good performance. 
The encoder module mutually refines the features of the sampled points and the key points. Subsequently, the outputs of the encoder module are fed into a Transformer decoder to generate a global representation. 


\subsection{Channel-wise Decoder and Training Losses}\label{sec.ct3d++_decoder_and_loss}
After obtaining the enhanced point features through the PBC scheme, it is necessary to aggregate them into a single global feature representation. To accomplish this, we employ the channel-wise decoder mentioned in Section \ref{sec.ct3d_decoder} to summarize the point features, represented as $\mathbf{F}^{update}\in \mathbb{R}^{N\times D}$, into a compact form denoted as $\mathbf{F}^{dec}\in\mathbb{R}^{1\times D}$. Subsequently, two FFNs are utilized to generate the final regression residuals and confidence estimation. The final confidence score is derived by averaging the outputs from both the RPN and the second-stage estimations. Noting that CT3D++ employs the same training loss approach as that of CT3D.

\section{Experiments}\label{sec:experiments}
In this section, we provide exhaustive experimental settings and implementation details for both the CT3D and CT3D++ frameworks. Subsequently, we compare our proposed CT3D and CT3D++ methods with state-of-the-art point cloud-based 3D object detection approaches. Finally, we conduct comprehensive ablation studies to assess the effectiveness of each component in both CT3D and CT3D++ methods.
\begin{table*}[t]  
\setlength\tabcolsep{1pt}
 	\caption{Performance comparisons with state-of-the-art methods on the validation set of Waymo Open Dataset for vehicle detection.} 
   \begin{center}
		{	\begin{tabular}{l||rrrr|rrrr}
				\hline
				\multirow{2}*{Method} & \multicolumn{4}{c|}{3D AP/APH (IoU=0.7)} &  \multicolumn{4}{c}{BEV AP/APH (IoU=0.7)}\\ 
				& Overall & 0-30m & 30-50m & 50m-Inf & Overall & 0-30m & 30-50m & 50m-Inf\\
				\hline
				{\textit{\textbf{LEVEL\_1}}}&&&&&&&\\
				PV-RCNN \cite{shi2020pv}  & 70.3/69.7 & 91.9/91.3 & 69.2/68.5 & 42.2/41.3 &80.0/82.1& 97.4/96.7& 83.0/82.0 &65.0/63.2\\
				Voxel-RCNN \cite{deng2021voxel}  & 75.6/- & 92.5/- & 74.1/- & 53.2/-&88.2/-& 97.6/-& 87.3/-& 77.7/-\\
				LiDAR-RCNN \cite{li2021lidar} &76.0/75.5& 92.1/91.6&74.6/74.1 &54.5/53.4&90.1/89.3&97.0/96.5&89.5/88.6& 78.9/77.4\\
				CenterPoint \cite{yin2021center}  &76.7/76.2&-&-&-&&-&-&-\\
				VoTr-TSD \cite{mao2021voxel} &75.0/74.3&92.3/91.7&73.4/72.6&51.1/50.0&-&-&-&-\\

                VoxelSet \cite{he2022voxel} & 77.8/- &92.8/-& 77.2/- &54.4/-&90.3/-& 96.1/-& 88.1/- &78.0/-\\

                RDIoU \cite{sheng2022rethinking}   &78.4/78.0&93.0/92.6&75.4/74.9&56.2/55.6&91.6/91.0&98.1/97.7&90.8/90.2&82.4/81.1\\
                IA-SSD\cite{zhang2022notall}  & 70.5/69.7&-&-&-&-&-&-&- \\
				BtcDet \cite{xu2022behind} &78.6/78.1&\textbf{96.1}/-&77.6/-&54.5/-&-&-&-&-\\

                AFDetV2 \cite{hu2022afdetv2} & 77.6/77.1&-&-&-&-&-&-&-\\
                
                PV-RCNN++ \cite{shi2023pv} &79.3/78.8&-&-&-&-&-&-&- \\
                LargeKernel3D \cite{chen2023largekernel3d} & 78.1/77.6 &-&-&-&-&-&-&- \\

                \hline
                \rowcolor{gray!40}CT3D&76.3/75.8&92.5/92.0& 75.1/74.6&55.4/54.5&90.5/90.0&97.6/97.0&88.1/87.1&78.9/77.4\\
                \rowcolor{gray!40}CT3D++&\textbf{80.5/79.9}&93.7/93.2&\textbf{79.3/78.8}&\textbf{60.0/59.3}&\textbf{92.2/91.5}&\textbf{98.2/97.8}&\textbf{91.3/90.7}&\textbf{83.6/82.4}\\
			\hline
                \hline
                {\textit{\textbf{LEVEL\_2}}}&&&&&&&\\
				PV-RCNN \cite{shi2020pv}  & 65.4/64.8 & 91.6/91.0 & 65.1/64.5 & 36.5/35.7 &77.5/76.6& 94.6/94.0& 80.4/79.4& 55.4/53.8\\
				Voxel-RCNN \cite{deng2021voxel}  & 66.6/- & 91.7/- & 67.9/- & 40.8/-&81.1/-& 97.0/-& 81.4/-& 63.3/-\\
				LiDAR-RCNN \cite{li2021lidar} &68.3/67.9&91.3/90.9& 68.5/68.0& 42.4/41.8&81.7/81.0&94.3/93.9&82.3/81.5&65.8/64.5\\
				CenterPoint \cite{yin2021center}  &68.8/68.3&-&-&-&-&-&-&-\\
				VoTr-TSD \cite{mao2021voxel} &65.9/65.3&-&-&-&-&-&-&-\\
                VoxelSet \cite{he2022voxel} &70.21/-& 92.05/-& 70.10/-& 43.20/-& 80.56/-& 96.79/-& 80.44/-& 62.37/-\\
                RDIoU \cite{sheng2022rethinking}  &69.5/69.1&92.3/91.9&69.3/68.9&43.7/43.1&83.1/82.5&97.5/97.1&85.2/84.6&68.3/67.2\\
				BtcDet \cite{xu2022behind} &70.1/69.6&\textbf{96.0/-}&70.1/-&43.9/-&-&-&-&-\\
                IA-SSD \cite{zhang2022notall}  & 61.6/60.8&-&-&-&-&-&-&- \\
                
                PV-RCNN++ \cite{shi2023pv} &70.6/70.2&-&-&-&-&-&-&-\\
                LargeKernel3D \cite{chen2023largekernel3d} & 69.8/69.4&-&-&-&-&-&-&-\\

                \hline
                \rowcolor{gray!40}CT3D&69.1/68.6&91.8/91.2&68.9/68.4&42.6/42.0&81.7/81.2&97.1/96.6&82.2/81.5&64.3/63.4\\
                \rowcolor{gray!40}CT3D++&\textbf{71.5/71.0}&93.0/92.5&\textbf{73.3/72.8}&\textbf{48.1/47.5}&\textbf{85.7/85.0}&\textbf{97.7/97.3}&\textbf{85.7/85.1}&\textbf{69.7/68.5}\\
				\hline
			\end{tabular}}
   \end{center}
	\label{table:waymo_val}
\end{table*}

\subsection{Datasets and Evaluation Metrics}

\subsubsection{Waymo Open Dataset} It collects 798 training and 202 validation sequences, where there are 158,361 LiDAR samples and 40,077 LiDAR samples, respectively. It has a much larger perception area supported by 5 LiDAR sensors as compared to the perception area of KITTI with only 1 LiDAR sensor. Currently, it is the largest outdoor 3D object detection dataset for autonomous driving. In main results, we report AP and average precision by heading (APH) performance on \textit{Vehicle} category with IoU larger than 0.7 for 3D/BEV detection. Moreover, performance on different distances (\textit{i.e.,}, 0-30m, 30-50m and 50m-\textit{Inf}) are separately exhibited, and two detection difficulty levels (\textit{i.e.,}, LEVEL\_1 with more than 5 points and LEVEL\_2 with no more than 5 points). In ablation studies, we also report \textit{Cyclist} results with IoU larger than 0.5.

\subsubsection{KITTI} It is a classical outdoor 3D object detection dataset for autonomous driving. It collects 7,481 frames of point cloud data for training and another 7,518 frames of point cloud data for testing. Following the common setting as in  \cite{yan2018second,lang2019pointpillars,shi2020pv,sheng2021improving}, we split the whole training data into \textit{train} set with 3,712 samples and \textit{val} set with 3,769 samples for experimental studies. We report the AP values for \textit{car} category with IoU larger than 0.7. In addition, we report \textit{Pedestrian} and \textit{Cyclist} categories with IoU larger than 0.5. Moreover, the results
on three difficulty levels \textit{i.e.,} \textit{easy}, \textit{moderate} and \textit{hard}, are exhibited according to the occlusion, truncation, and the number of corresponding image pixels of objects.

\subsection{Implementation Details}\label{implementation}
As shown in previous works  \cite{yan2018second,shi2020pv,deng2021voxel}, the implementation details for Waymo Open Dataset and KITTI are quite similar, and the major difference lies in the different sizes of perception area per frame. Our implementation mainly follows these previous works.
The default RPN consists of one 3D backbone and one 2D backbone. For the voxel size and XYZ ranges, we adopt the default setting in OpenPCDet \cite{openpcdet2020}.
The default 2D backbone contains two blocks, the first block is implemented by 5 CNN layers for KITTI and 6 CNN layers for Waymo Open Dataset to keep the same resolution with the output of 3D backbone. The second block is implemented by 1 CNN layer and 4 Transformer layers (using the setting of $4\times$ expansion in FFN layers and 4 attention heads) with half the resolution. Finally, one fractionally-strided convolution layer is adopted to double the resolution.
For KITTI, the default 3D backbone network is implemented by four-stage sparse convolution, and the filter numbers are 16, 32, 48, and 64, respectively. The fixed perception radius is 2.6m for \textit{car}, and 1.2m for both \textit{pedestrian} and \textit{cyclist}. For Waymo Open Dataset, the corresponding filter numbers are 16, 32, 64, and 128, respectively.  The default 2D backbone network is composed of 2d CNN and Swin-Transformer \cite{liu2021swin} layers. 
For CT3D, we set the perception radius scaling parameters $\alpha=1.2$. In object-based raw point sampling, we set $N=256$. For CT3D++, the fixed perception radius is 3.1m, 1.3m for \textit{vehicle}, \textit{pedestrian}/\textit{cyclist}, respectively.  In category-based raw point sampling, we set $N=255$.  For both object-based  and category-based raw point sampling strategies, we pad the sampled points with the first point if the number of points within the perception area is less than 255. 

\subsubsection{Training and Inference} All models are end-to-end trained from scratch. We use 8 V100 GPUs with batch size 24 for KITTI and batch size 16 for Waymo Open Dataset. CT3D's learning rate is $1e-3$, while CT3D++ uses $5e-4$. In training, 64 positive proposals are selected for regression and, with 64 negative samples, for confidence estimation. During inference, the top 100 3D proposals by classification score undergo further regression refinement and confidence evaluation.

\begin{table*}[t]  
 	\caption{Comparison with the state-of-the-art methods on the KITTI \textit{test} benchmark for \textit{car} detection.}
	\setlength\tabcolsep{8pt}
    
	\begin{center}
		{
			{	\begin{tabular}{l||rrrr|rrrr}
		\hline
		\multirow{2}*{Method}&\multicolumn{4}{c|}{Car-3D (IoU=0.7)} & \multicolumn{4}{c}{Car-BEV (IoU=0.7)}\\
		& mAP & Easy & Moderate & Hard  & mAP & Easy & Moderate &Hard\\
            \hline

        SA-SSD \cite{he2020structure}&80.90&88.75&79.79&74.16&90.67& 95.03&91.03&85.96\\
        PV-RCNN \cite{shi2020pv}&82.83&90.25& 81.43&76.82&90.59&94.98&90.65& 86.14\\
        Voxel R-CNN \cite{deng2021voxel}&83.19&90.90 &	81.62 &	77.06 &89.94&94.85 &88.83 &	86.13\\
        LiDAR-RCNN \cite{li2021lidar}&76.45&85.97&74.21&69.18&-&-&-&-\\
        VoTR-TSD \cite{mao2021voxel}&83.71&89.90 &	82.09 &	79.14&90.17&94.03 &	90.34 &	86.14\\
        VoxSeT \cite{he2022voxel}&82.68&88.53 &	82.06 &	77.46&89.35&92.70 &	89.07 &	86.29\\
        RDIoU \cite{sheng2022rethinking}&83.40&90.65 &	82.30 &	77.26&89.77&94.90 &	89.75 &	84.67\\
        BtcDet \cite{xu2022behind}&83.86&90.64 &	\textbf{82.86} &	78.09&88.90&92.81 &	89.34 &	84.55\\
        IA-SSD\cite{zhang2022notall} & 81.17 & 88.34 &	80.13 &	75.04  & 88.82 & 92.79 &	89.33 &	84.35 \\
        PV-RCNN++ \cite{shi2023pv}&83.06&90.14 &	81.88 &	77.15&89.12&92.66 &	88.74 &	85.97\\
        \hline
        \rowcolor{gray!40}CT3D &82.25&87.83 &	81.77 &	77.16&88.42&92.36 &	88.83 &	84.07\\
        \rowcolor{gray!40}CT3D++ &\textbf{84.25} &\textbf{90.91}&82.44&\textbf{79.41}&\textbf{91.66}&\textbf{95.12}&\textbf{91.35}&\textbf{88.50}\\
	
	\hline
	\end{tabular}}
		}
	\end{center}
    \label{table:kitti_test}
\end{table*}

\begin{table*}[tp]  
 	\caption{Performance comparisons with state-of-the-art methods on the KITTI \textit{val} set with 11 recall positions. All models reported here are trained with all the three categories.}
        \setlength\tabcolsep{1pt}
    
	\begin{center}
		{
			\scalebox{1.0}{	\begin{tabular}{l||rrrr|rrrr|rrrr}
		\hline
		\multirow{2}*{Method}& \multicolumn{4}{c|}{Car-3D (IoU=0.7)} & \multicolumn{4}{c|}{Ped.-3D (IoU=0.5)} & \multicolumn{4}{c}{Cyc.-3D (IoU=0.5)}\\
			& mAP & Easy & Moderate & Hard & mAP & Easy & Moderate &Hard & mAP & Easy & Moderate &Hard\\
            \hline
			VoxelNet \cite{zhou2018voxelnet} & 70.09 &81.97 &65.46 &62.85 &53.38&57.86& 53.42& 48.87&53.31& 67.17 &47.65 &45.11\\
	        SECOND \cite{yan2018second} &81.48&88.61& 78.62&77.22 &52.42&56.55&52.98&47.73&70.29&80.59&67.16&63.11\\
			PointPillar \cite{lang2019pointpillars}&79.46&86.46&77.28&74.65&52.65&57.75&52.29&47.91&67.49&80.06&62.69& 59.71\\
			3DSSD \cite{yang20203dssd}&82.61&\textbf{89.71}&79.45& 78.67&-&-&-&-&-&-&-&-\\
    
		Part-$A^2$ \cite{shi2020points}&82.60&89.56&79.41&78.84&60.40&65.69& 60.05&55.45&73.63&85.50&69.90&65.49\\
	   PV-RCNN \cite{shi2020pv} &83.91&89.35& 83.69& 78.70&56.58&63.12&54.84&51.78&73.35& 86.06&69.48&64.50\\
        RDIoU \cite{sheng2022rethinking}&84.27&89.16&85.24&78.41&57.75&63.26&57.47&52.53&71.78&83.32&68.39&63.63\\
        \hline
        \rowcolor{gray!40}CT3D&84.30&89.11& 85.04& 78.76&59.94&64.23& 59.84& 55.76&74.93&85.04&71.71&\textbf{68.05} \\
        \rowcolor{gray!40}CT3D++ & \textbf{86.77}&89.53&\textbf{85.28}&\textbf{85.51}&\textbf{63.06}&\textbf{67.89}&\textbf{63.24}&\textbf{58.04}&\textbf{75.55}&\textbf{86.33}&\textbf{72.45}&67.87\\
	
	\hline
	\end{tabular}}
		}
	\end{center}
    \label{table:kitti_val}
\end{table*}

\subsection{Comparison Results on Waymo}
We train our model on the entire Waymo training set and evaluate its performance on the Waymo validation set, as presented in Table \ref{table:waymo_val}. Our CT3D++ method outperforms previous state-of-the-art works, such as RDIoU \cite{sheng2022rethinking}, BtcDet \cite{xu2022behind}, and PV-RCNN++ \cite{shi2023pv}, in both 3D and BEV detection tasks, achieving remarkable performance gains. Specifically, our proposed method achieves a significant improvement of +1.9\%, +1.4\%, and +0.8\% in terms of APH of LEVEL\_2 difficulty for 3D detection, which is the most important metric for Waymo Open Dataset, compared to RDIoU, BtcDet, and PV-RCNN++, respectively. Moreover, for BEV detection, our CT3D++ surpasses all other methods in all distance ranges and difficulty levels by large margins. Notably, our method brings more improvement for objects with distances greater than 30m, which often suffer from occlusion and signal miss problems and thus require semantic information.  As compared to our previous proposed method CT3D, CT3D++ surpasses it with 2.4\%APH. In conclusion, our consistent performance improvements on the largest point cloud-based 3D object detection dataset demonstrate the effectiveness of our proposed CT3D++ method. 

\subsection{Comparison Results on KITTI}
We train the model on all the KITTI training data (\textit{i.e.,}, \textit{trainval} set), and report the \textit{car} category detection performance of our method on KITTI \textit{test} benchmark. The comparison results with the state-of-the-art methods are listed in Table \ref{table:kitti_test}. Our proposed CT3D++ method ranks in the $1^{st}$ place over almost all the metrics with significant improvement.
Specifically, we achieve $90.91\%$, $82.44\%$ and $79.41\%$ AP on 3D detection for \textit{car} category with $84.25\%$ mean average precision (mAP). Our method outperforms the most recent work RDIoU \cite{sheng2021improving}, BtcDet \cite{xu2022behind} with $+0.85\%$ and $+0.39\%$ improvement, respectively, on 3D detection. Additionally, our method surpasses RDIoU and BtcDet with $+1.89\%$ and $+2.76\%$ improvement, respectively, on BEV detection, which is also an important task when evaluating on KITTI \textit{test} benchmark \cite{geiger2013vision}. 


In addition, we also provide the comparisons with state-of-the-art methods in terms of three categories on KITTI \textit{val} set in Table \ref{table:kitti_val}. Unlike many related works \cite{deng2021voxel,he2020structure,mao2021voxel} that only train models on \textit{car} category or train different models for different categories \cite{xu2022behind}, our proposed CT3D++ conducts multi-class training and still can achieve promising performance. Particularly, our CT3D++ leads a large margin as compared to the second-best method CT3D \cite{sheng2021improving}. For \textit{car} detection, we can outperform the current best performance with $+2.47\%$ on mAP. Especially, our CT3D++ significantly improves the detection performance on \textit{pedestrian} and \textit{cyclist} categories, leading to $+2.66\%$ and $+0.62\%$ improvements on mAP, respectively, compared with the second-best performance. To sum up, our CT3D++ performs significantly better than state-of-the-art methods over almost all categories and all difficulty levels. There only exists tiny performance drop on \textit{easy}-level \textit{car} detection and \textit{hard}-level \textit{cyclist} detection. These strong results further verify the effectiveness of our method.

\subsection{Ablation Study}\label{sec:abla}
We conduct extensive experiments to analyze the effects of major contributions in our method. All models are trained on $20\%$ Waymo \textit{train} set like PV-RCNN++ \cite{shi2023pv} and evaluated on full Waymo \textit{val} set. During training, we fix the RPN parameters for fare comparisons and train the second-stage refinement network with the fixed learning rate (\textit{i.e.,}, $5e-4$), training epochs (\textit{i.e.,}, 16) and batch size (\textit{i.e.,}, 24). We report the performance of \textit{Vehicle} and \textit{Cyclist} detection on APH metric with LEVEL\_1 and LEVEL\_2, respectively.

\subsubsection{Effects on different RPNs}
To demonstrate the flexibility of our CT3D++ framework, we conduct experiments using two different RPNs: PointPillar \cite{lang2019pointpillars} and CT-stacked \cite{sheng2022rethinking}. The results, reported in Table \ref{table:abla_rpn}, illustrate that our method consistently outperforms both RPNs with significant margins across all evaluation metrics.

\subsubsection{Effects on channel-wise decoder}
Table \ref{table:abla_channel_wise_decoder} demonstrates that the extended channel-wise re-wei\-ghting significantly outperforms both the standard cross-att\-ention and channel-wise re-weighting approaches. This substantial improvement can be attributed to the integration of standard decoding and channel-wise re-weighting techniques, which enables more effective decoding weights through global and channel-wise local aggregation.


\begin{figure*}[t]
	\begin{center}
		\includegraphics[width=0.95\linewidth]{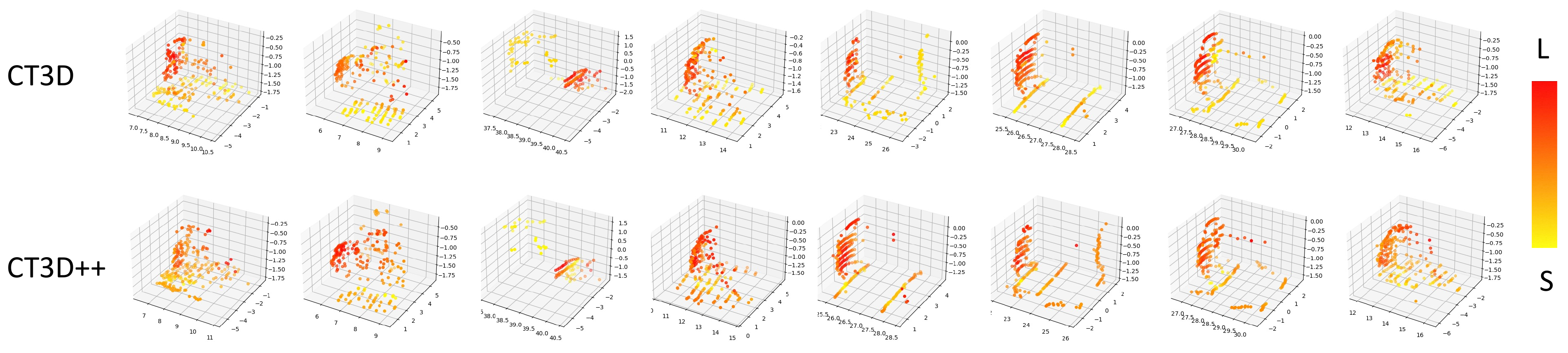}
	\end{center}
	\caption{Attention maps of several instances generated by encoder of CT3D and CT3D++, respectively.}
	\label{fig:attention}
\end{figure*}

\begin{table}[t]
    \caption{Ablation study on different RPNs.} \label{table:abla_rpn}
    \setlength\tabcolsep{4pt}
    {
    \begin{center}
	\begin{tabular}{l||rr|rr}
	\hline
	\multirow{2}*{Method}&\multicolumn{2}{c|}{Vehicle} & \multicolumn{2}{c}{Cyclist} \\
	&  L\_1 & L\_2 & L\_1 & L\_2 \\
	\hline
        PointPillar \cite{lang2019pointpillars} & 69.7 & 61.5 & 51.8 & 49.9\\
        \rowcolor{gray!40}CT3D (PointPillar) & 70.2 & 62.6 & 54.4 & 52.3 \\
        \rowcolor{gray!40}CT3D++ (PointPillar) & 73.8 & 64.8 & 58.5 & 56.2\\
        CT-stacked \cite{sheng2022rethinking} & 74.3 & 65.4 & 54.4 & 52.6\\
        \rowcolor{gray!40}CT3D (CT-stacked) & 75.1 & 66.7 & 57.9 & 55.7\\
        \rowcolor{gray!40}CT3D++ (CT-stacked) & 76.3 & 68.9 & 63.6 & 61.1\\
	\hline
	\end{tabular}
    \end{center}
    }
\end{table}

\begin{table}[t]
    \caption{Ablation study on channel-wise decoder.}\label{table:abla_channel_wise_decoder} 
    \setlength\tabcolsep{2.5pt}
    {
    \begin{center}
	\begin{tabular}{l||rr|rr}
	\hline
	\multirow{2}*{Method}&\multicolumn{2}{c|}{Vehicle} & \multicolumn{2}{c}{Cyclist} \\
	&  L\_1 & L\_2 & L\_1 & L\_2 \\
	\hline
        Standard cross-attention  & 73.9 &  65.2 & 55.5 & 53.5\\
        Channel-wise re-weighting & 73.9 &  65.3 & 55.5 & 53.6\\
        \hline
        \rowcolor{gray!40}Extended channel-wise re-weighting & 76.3 & 68.9 & 63.6 & 61.1\\
	\hline
	\end{tabular}
    \end{center}
    }
\end{table}

\begin{table}[t]
    \caption{Ablation study on different refinement networks. We reproduce (I): Voxel R-CNN \cite{deng2021voxel} (III): PV-RCNN \cite{shi2020pv} and (IV): PV-RCNN++ \cite{shi2023pv} with same RPN. (IV): CT3D (V): CT3D++
    Time is tested by 1 V100 GPU with batch size 1.}\label{table:abla_refinement} 
    \setlength\tabcolsep{5pt}
    {
    \begin{center}
	\begin{tabular}{l|r|r||rr|rr}
	\hline
	\multirow{2}*{Method}& \multirow{2}*{Refine}&{Time} &\multicolumn{2}{c|}{Vehicle} & \multicolumn{2}{c}{Cyclist} \\
	&& (ms)& L\_1 & L\_2 & L\_1 & L\_2 \\
	\hline
        Voxel R-CNN &SIR&85.6& 74.3 & 65.5 & 62.0 & 59.6\\
        PV-RCNN  & HIR &367.5& 73.9 & 65.1 & 56.9 & 54.9\\
        PV-RCNN++  & HIR &225.9& 75.2 & 67.5 & 61.0 & 58.7\\
        \hline
        \rowcolor{gray!40} CT3D &GIR&178.6& 73.0 & 64.3 & 55.1 & 53.0\\
        \rowcolor{gray!40} CT3D++ &HIR& 146.2 & 76.3 & 68.9 & 63.6 & 61.1\\
	\hline
	\end{tabular}
    \end{center}
    }

\end{table}

\begin{table}[t]
    \caption{Ablation study on the geometric and semantic fusion embedding. Vox. means incorporating voxel features extracted from the middle layers of RPN.}\label{table:abla_embedding}
    {
    \begin{center}
	\begin{tabular}{cll||ll|ll}
	\hline
	\multirow{2}*{Geo.} & \multirow{2}*{Sem.} & \multirow{2}*{Vox.} & \multicolumn{2}{c|}{Vehicle} & \multicolumn{2}{c}{Cyclist} \\
	&&& L\_1 & L\_2 & L\_1 & L\_2\\
	\hline
        \checkmark &&& 76.1 & 68.2 & 63.4 & 60.5\\
        & \checkmark && 73.0 & 64.1 & 53.8 & 51.9\\
        \checkmark & \checkmark & \checkmark & 76.3 & 69.0 & 63.6 & 61.1\\
        \hline
        \rowcolor{gray!40} \checkmark & \checkmark & & 76.3 & 68.9 & 63.6 & 61.1\\
	\hline
	\end{tabular}
    \end{center}
    }
\end{table}

\begin{figure}[t]
	\begin{center}
		\includegraphics[width=0.7\linewidth]{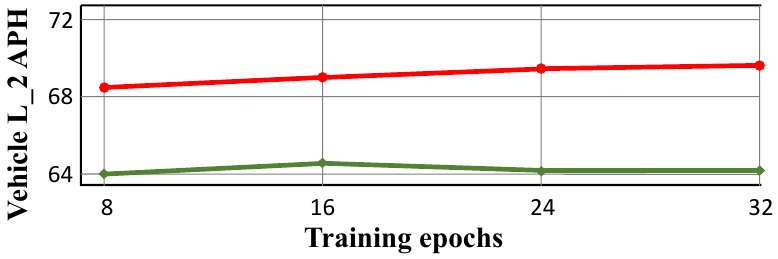}
	\end{center}
	\caption{Performance comparison of self-attention and our proposed PBC schemes using \textit{Vehicle} LEVEL\_2 APH metric.}
	\label{fig:self-pbc}
\end{figure}

\begin{table}[t]
    %
    \caption{Ablation study on our proposed PBC scheme.}\label{table:abla_pbc}
    {
    \begin{center}
	\begin{tabular}{l||r|rr|rr}
	\hline
	\multirow{2}*{Encoding scheme} & Time & \multicolumn{2}{c|}{Vehicle} & \multicolumn{2}{c}{Cyclist} \\
	& (ms) &  L\_1 & L\_2 & L\_1 & L\_2\\
	\hline
        MLP & 1.3 & 75.4 & 66.4 & 62.7 & 60.2\\
        Self-attention & 23.9 & 73.4 & 64.5 & 55.3 & 53.6\\
        \hline
        \rowcolor{gray!40} PBC & 16.6 & 76.3 & 68.9 & 63.6 & 61.1\\
	\hline
	\end{tabular}
    \end{center}
    }
\end{table}

\subsubsection{Effects on different refinement networks} 
We compare with four other refinement methods, including one SIR, one GIR, and two HIR approaches. To ensure a fair comparison, we use the same fixed-parameter RPN for all four methods, namely Voxel R-CNN \cite{deng2021voxel}, CT3D \cite{sheng2021improving}, PV-RCNN \cite{shi2020pv}, and PV-RCNN++\cite{shi2023pv}. The results are presented in Table \ref{table:abla_refinement}, showing that our method significantly outperforms all the other methods. We attribute our success to the fusion of geometric and semantic information, which is absent in Voxel R-CNN and CT3D. Although PV-RCNN and PV-RCNN++ are also HIR methods that integrate both types of information, their performance falls short of ours, further demonstrating the efficacy of our fusion-based refinement network. Regarding latency, our CT3D++ requires only 146.2ms per frame when run on a single V100 GPU, while achieving significantly improved performance as compared to CT3D \cite{sheng2021improving}, PV-RCNN \cite{shi2020pv} and PV-RCNN++ \cite{shi2023pv} with much less time required.

\begin{table}[t]
    \caption{Ablation study on different kinds of decoding points.}\label{table:abla_rpn_de} 
    {
    \begin{center}
	\begin{tabular}{l||rr|rr}
	\hline
	\multirow{2}*{Decoding on}&\multicolumn{2}{c|}{Vehicle} & \multicolumn{2}{c}{Cyclist} \\
	&  L\_1 & L\_2 & L\_1 & L\_2 \\
	\hline
        Key points & 76.0 & 68.6 & 62.9 & 60.4\\
        \rowcolor{gray!40}Raw points & 76.3 & 68.9 & 63.6 & 61.1\\
	\hline
	\end{tabular}
    \end{center}
    }
\end{table}

\begin{table}[t]
    \caption{Ablation study on the different sampling strategies. Time is tested with 1 V100 GPU and batch size 4.}\label{table:abla_sampling} 
    {
    \begin{center}
	\begin{tabular}{l|r||rr|rr}
	\hline
	\multirow{2}*{Sampling}& Time &\multicolumn{2}{c|}{Vehicle} & \multicolumn{2}{c}{Cyclist} \\
	& (ms) &  L\_1 & L\_2 & L\_1 & L\_2 \\
	\hline
        Object-based & 111.4 & 76.4 & 69.0 & 63.8 & 61.2\\
        \rowcolor{gray!40} Category-based & 60.5 & 76.3 & 68.9 & 63.6 & 61.1\\
	\hline
	\end{tabular}
    \end{center}
    }
\end{table}

\begin{table}[ht]
    \caption{Comparisons between the standard self-attention  and our proposed PBC schemes with different training epochs. The performance of the last epoch is reported.}\label{table:abla_pbc2} 
    \setlength\tabcolsep{4.5pt}%
    {
    \begin{center}
	\begin{tabular}{l|c||cc|cc}
	\hline
	\multirow{2}*{Encoding scheme}&Training&\multicolumn{2}{c|}{Vehicle} & \multicolumn{2}{c}{Cyclist} \\
	& epochs & L\_1 & L\_2 & L\_1 & L\_2 \\
	\hline
        Self-attention & 8 & 72.8 & 64.0 & 54.3 & 52.4\\
        \rowcolor{gray!40}PBC & 8 & 75.9 & 68.5 & 62.8 & 60.2 \\
        \hline
        Self-attention & 16 & 73.4 & 64.5 & 55.3 & 53.6\\
        \rowcolor{gray!40}PBC & 16 & 76.3 & 68.9 & 63.6 & 61.1\\
        \hline
        Self-attention & 24 & 72.9 & 64.1 & 55.6 & 53.6\\
        \rowcolor{gray!40}PBC & 24 & 76.7 & 69.3 & 64.5 & 62.3\\
        \hline
        Self-attention & 32 & 72.9 & 64.1 & 55.7 & 53.5\\
        \rowcolor{gray!40}PBC & 32 & 78.5 & 69.4 & 64.8 & 62.6\\
	\hline
	\end{tabular}
    \end{center}
    }

\end{table}

\subsubsection{Effects on geometric and semantic fusion embedding} We evaluate the effectiveness of this module by adding different features from RPN, as shown in Table \ref{table:abla_embedding}. The results demonstrate that the combination of geometric features (\textit{i.e.,}, raw points) and semantic features (\textit{i.e.,}, BEV feature map) leads to significant performance improvements. We also test adding voxel features from the middle layers of RPN, similar to Voxel R-CNN \cite{deng2021voxel}, and observe no significant improvement. This confirms that the semantic information provided by BEV features is sufficient.

\subsubsection{Effects on the PBC scheme} We compare our proposed scheme with other feature enhancement techniques, such as the MLP and self-attention \cite{vaswani2017attention} in Table \ref{table:abla_pbc}. Our PBC scheme exhibits significant better performance than both the MLP and self-attention schemes. This is because the self-attention scheme is highly complex and difficult to converge, while the capabilities of MLP networks are limited. The self-attention scheme is highly complex and difficult to converge, as evidenced by its poorer performance compared to the simple MLP, as shown in the $1^{st}$ and $2^{nd}$ rows of the table. In contrast, our PBC scheme explicitly models the relationship between the raw points and key points, making it more efficient in estimating residuals and confidence. Moreover, our PBC scheme achieves a 31.3\% reduction in runtime compared to the self-attention scheme, while still significantly improving 3D detection performance. This indicates that our method is highly efficient. Additionally, in Figure \ref{fig:attention}, we visualize the attention maps of several instances generated by our PBC scheme, which clearly shows that the network assigns more attention (red) to foreground points and less attention (yellow) to background points.

\subsubsection{Effects on the different kinds of decoding points} Table \ref{table:abla_rpn_de} shows the results of experiments conducted to investigate the influence of different kinds of  decoding points (\textit{i.e.,}, 255 sampled raw points vs. 9 key points). While there is information exchange between the raw point features and key point features, the results show that decoding with raw points outperforms key points-based decoding.

\subsubsection{Effects on the category-based raw point sampling} Table \ref{table:abla_sampling} compares our category-based raw point sampling strategy with object-based raw point sampling. The results indicate that category-based raw point sampling can significantly increase the training/inference speed while only causing a slight performance drop compared to object-based raw point sampling. These results demonstrate that our method is not highly sensitive to the sampled range and can maintain robustness to incomplete point clouds.

\subsubsection{More comparisons between PBC and self-attention} Table \ref{table:abla_pbc2} presents additional experiments conducted to validate the effectiveness of our proposed PBC scheme. The findings indicate that our method attains accelerated convergence and superior performance relative to the conventional self-attention framework, thereby furnishing additional corroboration of the efficacy of our PBC scheme. In addition, we plot the corresponding line chart of \textit{Vehicle} LEVEL\_2 APH in Figure \ref{fig:self-pbc}, which clearly shows that our PBC scheme outperforms the standard self-attention scheme in terms of convergence and performance.

\section{Conclusion}
In this paper, we have presented two innovative frameworks, CT3D and CT3D++, which significantly advance the state-of-the-art in 3D object detection from point clouds. Our work is predicated on the idea that the key to improving detection performance lies in minimal hand-crafted design and in leveraging the strengths of Transformers for feature encoding. CT3D lays the foundation by introducing a raw-point-based embedding process followed by a standard Transformer encoder and a channel-wise decoder, which together enhance the process of feature extraction from point clouds. Building upon the success of CT3D, we developed CT3D++, which further refines the detection process through the integration of geometric and semantic information. This is achieved by a geometric and semantic fusion-based embedding technique that enriches proposal-aware features with more meaningful context. Moreover, the innovative point-to-key bidirectional encoder implemented in CT3D++ not only improves feature representation but also reduces computational overhead, making it a more efficient and practical solution for real-world applications. Our experimental results demonstrate that both CT3D and CT3D++ achieve superior performance on benchmark datasets like KITTI and Waymo Open Dataset, confirming the effectiveness of our approaches in handling the complexity and variability of real-world 3D environments. The scalability and adaptability of our frameworks make them suitable for a wide array of applications in autonomous driving, robotics, and beyond.



\newpage
\bibliographystyle{plainnat}
\bibliography{ct3d++.bib}

\begin{thebibliography}{78}
\providecommand{\natexlab}[1]{#1}
\providecommand{\url}[1]{\texttt{#1}}
\expandafter\ifx\csname urlstyle\endcsname\relax
  \providecommand{\doi}[1]{doi: #1}\else
  \providecommand{\doi}{doi: \begingroup \urlstyle{rm}\Url}\fi

\bibitem[Cao et~al.(2017)Cao, Wu, and Shen]{cao2017estimating}
Yuanzhouhan Cao, Zifeng Wu, and Chunhua Shen.
\newblock Estimating depth from monocular images as classification using deep
  fully convolutional residual networks.
\newblock \emph{IEEE Transactions on Circuits and Systems for Video Technology
  (TCSVT)}, 28\penalty0 (11):\penalty0 3174--3182, 2017.

\bibitem[Carion et~al.(2020)Carion, Massa, Synnaeve, Usunier, Kirillov, and
  Zagoruyko]{carion2020end}
Nicolas Carion, Francisco Massa, Gabriel Synnaeve, Nicolas Usunier, Alexander
  Kirillov, and Sergey Zagoruyko.
\newblock End-to-end object detection with transformers.
\newblock In \emph{Proceedings of the European Conference on Computer Vision},
  pages 213--229. Springer, 2020.

\bibitem[Chai et~al.(2021)Chai, Sun, Ngiam, Wang, Caine, Vasudevan, Zhang, and
  Anguelov]{chai2021point}
Yuning Chai, Pei Sun, Jiquan Ngiam, Weiyue Wang, Benjamin Caine, Vijay
  Vasudevan, Xiao Zhang, and Dragomir Anguelov.
\newblock To the point: Efficient 3d object detection in the range image with
  graph convolution kernels.
\newblock In \emph{Proceedings of the IEEE/CVF Conference on Computer Vision
  and Pattern Recognition}, pages 16000--16009, 2021.

\bibitem[Chen et~al.(2021)Chen, Pu, Fan, and Zou]{chen2021fixing}
Shu Chen, Zhengdong Pu, Xiang Fan, and Beiji Zou.
\newblock Fixing defect of photometric loss for self-supervised monocular depth
  estimation.
\newblock \emph{IEEE Transactions on Circuits and Systems for Video Technology
  (TCSVT)}, 32\penalty0 (3):\penalty0 1328--1338, 2021.

\bibitem[Chen et~al.(2017)Chen, Ma, Wan, Li, and Xia]{chen2017multi}
Xiaozhi Chen, Huimin Ma, Ji~Wan, Bo~Li, and Tian Xia.
\newblock Multi-view 3d object detection network for autonomous driving.
\newblock In \emph{Proceedings of the IEEE Conference on Computer Vision and
  Pattern Recognition (CVPR)}, pages 1907--1915, 2017.

\bibitem[Chen et~al.(2020)Chen, Tai, Sun, and Li]{chen2020monopair}
Yongjian Chen, Lei Tai, Kai Sun, and Mingyang Li.
\newblock Monopair: Monocular 3d object detection using pairwise spatial
  relationships.
\newblock In \emph{Proceedings of the IEEE/CVF Conference on Computer Vision
  and Pattern Recognition (CVPR)}, pages 12093--12102, 2020.

\bibitem[Chen et~al.(2023)Chen, Liu, Zhang, Qi, and Jia]{chen2023largekernel3d}
Yukang Chen, Jianhui Liu, Xiangyu Zhang, Xiaojuan Qi, and Jiaya Jia.
\newblock Largekernel3d: Scaling up kernels in 3d sparse cnns.
\newblock In \emph{Proceedings of the IEEE/CVF Conference on Computer Vision
  and Pattern Recognition}, pages 13488--13498, 2023.

\bibitem[Deng et~al.(2020)Deng, Shi, Li, Zhou, Zhang, and Li]{deng2020voxel}
Jiajun Deng, Shaoshuai Shi, Peiwei Li, Wengang Zhou, Yanyong Zhang, and
  Houqiang Li.
\newblock Voxel r-cnn: Towards high performance voxel-based 3d object
  detection.
\newblock \emph{arXiv preprint arXiv:2012.15712}, 2020.

\bibitem[Deng et~al.(2021)Deng, Shi, Li, Zhou, Zhang, and Li]{deng2021voxel}
Jiajun Deng, Shaoshuai Shi, Peiwei Li, Wengang Zhou, Yanyong Zhang, and
  Houqiang Li.
\newblock Voxel r-cnn: Towards high performance voxel-based 3d object
  detection.
\newblock \emph{Proceedings of the AAAI Conference on Artificial Intelligence},
  35\penalty0 (2):\penalty0 1201--1209, 2021.

\bibitem[Ding et~al.(2020)Ding, Huo, Yi, Wang, Shi, Lu, and
  Luo]{ding2020learning}
Mingyu Ding, Yuqi Huo, Hongwei Yi, Zhe Wang, Jianping Shi, Zhiwu Lu, and Ping
  Luo.
\newblock Learning depth-guided convolutions for monocular 3d object detection.
\newblock In \emph{Proceedings of the IEEE/CVF Conference on Computer Vision
  and Pattern Recognition Workshops}, pages 1000--1001, 2020.

\bibitem[Dosovitskiy et~al.(2020)Dosovitskiy, Beyer, Kolesnikov, Weissenborn,
  Zhai, Unterthiner, Dehghani, Minderer, Heigold, Gelly,
  et~al.]{dosovitskiy2020image}
Alexey Dosovitskiy, Lucas Beyer, Alexander Kolesnikov, Dirk Weissenborn,
  Xiaohua Zhai, Thomas Unterthiner, Mostafa Dehghani, Matthias Minderer, Georg
  Heigold, Sylvain Gelly, et~al.
\newblock An image is worth 16x16 words: Transformers for image recognition at
  scale.
\newblock \emph{arXiv preprint arXiv:2010.11929}, 2020.

\bibitem[Fan et~al.(2021)Fan, Xiong, Wang, Wang, and Zhang]{fan2021rangedet}
Lue Fan, Xuan Xiong, Feng Wang, Naiyan Wang, and Zhaoxiang Zhang.
\newblock Rangedet: In defense of range view for lidar-based 3d object
  detection.
\newblock In \emph{Proceedings of the IEEE/CVF International Conference on
  Computer Vision}, pages 2918--2927, 2021.

\bibitem[Geiger et~al.(2013)Geiger, Lenz, Stiller, and
  Urtasun]{geiger2013vision}
Andreas Geiger, Philip Lenz, Christoph Stiller, and Raquel Urtasun.
\newblock Vision meets robotics: The kitti dataset.
\newblock \emph{The International Journal of Robotics Research}, 32\penalty0
  (11):\penalty0 1231--1237, 2013.

\bibitem[Guo et~al.(2021)Guo, Cai, Liu, Mu, Martin, and Hu]{guo2021pct}
Meng-Hao Guo, Jun-Xiong Cai, Zheng-Ning Liu, Tai-Jiang Mu, Ralph~R Martin, and
  Shi-Min Hu.
\newblock Pct: Point cloud transformer.
\newblock \emph{Computational Visual Media}, 7\penalty0 (2):\penalty0 187--199,
  2021.

\bibitem[He et~al.(2020)He, Zeng, Huang, Hua, and Zhang]{he2020structure}
Chenhang He, Hui Zeng, Jianqiang Huang, Xian-Sheng Hua, and Lei Zhang.
\newblock Structure aware single-stage 3d object detection from point cloud.
\newblock In \emph{Proceedings of the IEEE Conference on Computer Vision and
  Pattern Recognition (CVPR)}, pages 11873--11882, 2020.

\bibitem[He et~al.(2022)He, Li, Li, and Zhang]{he2022voxel}
Chenhang He, Ruihuang Li, Shuai Li, and Lei Zhang.
\newblock Voxel set transformer: A set-to-set approach to 3d object detection
  from point clouds.
\newblock In \emph{Proceedings of the IEEE/CVF Conference on Computer Vision
  and Pattern Recognition}, pages 8417--8427, 2022.

\bibitem[Hu et~al.(2022)Hu, Ding, Ge, Shao, Huang, Li, and Liu]{hu2022afdetv2}
Yihan Hu, Zhuangzhuang Ding, Runzhou Ge, Wenxin Shao, Li~Huang, Kun Li, and
  Qiang Liu.
\newblock Afdetv2: Rethinking the necessity of the second stage for object
  detection from point clouds.
\newblock \emph{Proceedings of the AAAI Conference on Artificial Intelligence},
  36\penalty0 (1):\penalty0 969--979, 2022.

\bibitem[Huang and Huang(2022)]{huang2022bevdet4d}
Junjie Huang and Guan Huang.
\newblock Bevdet4d: Exploit temporal cues in multi-camera 3d object detection.
\newblock \emph{arXiv preprint arXiv:2203.17054}, 2022.

\bibitem[Huang et~al.(2021)Huang, Huang, Zhu, Ye, and Du]{huang2021bevdet}
Junjie Huang, Guan Huang, Zheng Zhu, Yun Ye, and Dalong Du.
\newblock Bevdet: High-performance multi-camera 3d object detection in
  bird-eye-view.
\newblock \emph{arXiv preprint arXiv:2112.11790}, 2021.

\bibitem[Ku et~al.(2018)Ku, Mozifian, Lee, Harakeh, and Waslander]{ku2018joint}
Jason Ku, Melissa Mozifian, Jungwook Lee, Ali Harakeh, and Steven~L Waslander.
\newblock Joint 3d proposal generation and object detection from view
  aggregation.
\newblock In \emph{2018 IEEE/RSJ International Conference on Intelligent Robots
  and Systems (IROS)}, pages 1--8, 2018.

\bibitem[Lang et~al.(2019)Lang, Vora, Caesar, Zhou, Yang, and
  Beijbom]{lang2019pointpillars}
Alex~H Lang, Sourabh Vora, Holger Caesar, Lubing Zhou, Jiong Yang, and Oscar
  Beijbom.
\newblock Pointpillars: Fast encoders for object detection from point clouds.
\newblock In \emph{Proceedings of the IEEE/CVF Conference on Computer Vision
  and Pattern Recognition (CVPR)}, pages 12697--12705, 2019.

\bibitem[Law and Deng(2018)]{law2018cornernet}
Hei Law and Jia Deng.
\newblock Cornernet: Detecting objects as paired keypoints.
\newblock In \emph{Proceedings of the European Conference on Computer Vision
  (ECCV)}, pages 734--750, 2018.

\bibitem[Lee et~al.(2019)Lee, Lee, Kim, Kosiorek, Choi, and Teh]{lee2019set}
Juho Lee, Yoonho Lee, Jungtaek Kim, Adam Kosiorek, Seungjin Choi, and Yee~Whye
  Teh.
\newblock Set transformer: A framework for attention-based
  permutation-invariant neural networks.
\newblock In \emph{International Conference on Machine Learning}, pages
  3744--3753. PMLR, 2019.

\bibitem[Li(2017)]{li20173d}
Bo~Li.
\newblock 3d fully convolutional network for vehicle detection in point cloud.
\newblock In \emph{2017 IEEE/RSJ International Conference on Intelligent Robots
  and Systems}, pages 1513--1518. IEEE, 2017.

\bibitem[Li et~al.(2023)Li, Ge, Yu, Yang, Wang, Shi, Sun, and
  Li]{li2023bevdepth}
Yinhao Li, Zheng Ge, Guanyi Yu, Jinrong Yang, Zengran Wang, Yukang Shi,
  Jianjian Sun, and Zeming Li.
\newblock Bevdepth: Acquisition of reliable depth for multi-view 3d object
  detection.
\newblock \emph{Proceedings of the AAAI Conference on Artificial Intelligence},
  37\penalty0 (2):\penalty0 1477--1485, 2023.

\bibitem[Li et~al.(2021)Li, Wang, and Wang]{li2021lidar}
Zhichao Li, Feng Wang, and Naiyan Wang.
\newblock Lidar r-cnn: An efficient and universal 3d object detector.
\newblock In \emph{Proceedings of the IEEE/CVF Conference on Computer Vision
  and Pattern Recognition}, pages 7546--7555, 2021.

\bibitem[Li et~al.(2022)Li, Wang, Li, Xie, Sima, Lu, Qiao, and
  Dai]{li2022bevformer}
Zhiqi Li, Wenhai Wang, Hongyang Li, Enze Xie, Chonghao Sima, Tong Lu, Yu~Qiao,
  and Jifeng Dai.
\newblock Bevformer: Learning bird’s-eye-view representation from
  multi-camera images via spatiotemporal transformers.
\newblock In \emph{European conference on computer vision}, pages 1--18.
  Springer, 2022.

\bibitem[Liang et~al.(2021)Liang, Zhang, Zhang, Zhao, and
  Pu]{liang2021rangeioudet}
Zhidong Liang, Zehan Zhang, Ming Zhang, Xian Zhao, and Shiliang Pu.
\newblock Rangeioudet: Range image based real-time 3d object detector optimized
  by intersection over union.
\newblock In \emph{Proceedings of the IEEE/CVF Conference on Computer Vision
  and Pattern Recognition}, pages 7140--7149, 2021.

\bibitem[Liu et~al.(2023{\natexlab{a}})Liu, Teng, Lu, Wang, and
  Wang]{liu2023sparsebev}
Haisong Liu, Yao Teng, Tao Lu, Haiguang Wang, and Limin Wang.
\newblock Sparsebev: High-performance sparse 3d object detection from
  multi-camera videos.
\newblock In \emph{Proceedings of the IEEE/CVF International Conference on
  Computer Vision}, pages 18580--18590, 2023{\natexlab{a}}.

\bibitem[Liu et~al.(2021{\natexlab{a}})Liu, Xue, and Wu]{liu2021learning}
Xianpeng Liu, Nan Xue, and Tianfu Wu.
\newblock Learning auxiliary monocular contexts helps monocular 3d object
  detection.
\newblock \emph{arXiv preprint arXiv:2112.04628}, 2021{\natexlab{a}}.

\bibitem[Liu et~al.(2022)Liu, Wang, Zhang, and Sun]{liu2022petr}
Yingfei Liu, Tiancai Wang, Xiangyu Zhang, and Jian Sun.
\newblock Petr: Position embedding transformation for multi-view 3d object
  detection.
\newblock In \emph{European Conference on Computer Vision}, pages 531--548.
  Springer, 2022.

\bibitem[Liu et~al.(2023{\natexlab{b}})Liu, Yan, Jia, Li, Gao, Wang, and
  Zhang]{liu2023petrv2}
Yingfei Liu, Junjie Yan, Fan Jia, Shuailin Li, Aqi Gao, Tiancai Wang, and
  Xiangyu Zhang.
\newblock Petrv2: A unified framework for 3d perception from multi-camera
  images.
\newblock In \emph{Proceedings of the IEEE/CVF International Conference on
  Computer Vision}, pages 3262--3272, 2023{\natexlab{b}}.

\bibitem[Liu et~al.(2021{\natexlab{b}})Liu, Yixuan, and Liu]{liu2021ground}
Yuxuan Liu, Yuan Yixuan, and Ming Liu.
\newblock Ground-aware monocular 3d object detection for autonomous driving.
\newblock \emph{IEEE Robotics and Automation Letters}, 6\penalty0 (2):\penalty0
  919--926, 2021{\natexlab{b}}.

\bibitem[Liu et~al.(2021{\natexlab{c}})Liu, Lin, Cao, Hu, Wei, Zhang, Lin, and
  Guo]{liu2021swin}
Ze~Liu, Yutong Lin, Yue Cao, Han Hu, Yixuan Wei, Zheng Zhang, Stephen Lin, and
  Baining Guo.
\newblock Swin transformer: Hierarchical vision transformer using shifted
  windows.
\newblock In \emph{Proceedings of the IEEE/CVF International Conference on
  Computer Vision}, pages 10012--10022, 2021{\natexlab{c}}.

\bibitem[Liu et~al.(2021{\natexlab{d}})Liu, Zhang, Cao, Hu, and
  Tong]{liu2021group}
Ze~Liu, Zheng Zhang, Yue Cao, Han Hu, and Xin Tong.
\newblock Group-free 3d object detection via transformers.
\newblock In \emph{Proceedings of the IEEE/CVF International Conference on
  Computer Vision}, pages 2949--2958, 2021{\natexlab{d}}.

\bibitem[Liu et~al.(2020)Liu, Wu, and T{\'o}th]{liu2020smoke}
Zechen Liu, Zizhang Wu, and Roland T{\'o}th.
\newblock Smoke: Single-stage monocular 3d object detection via keypoint
  estimation.
\newblock In \emph{Proceedings of the IEEE/CVF Conference on Computer Vision
  and Pattern Recognition Workshops}, pages 996--997, 2020.

\bibitem[Lu et~al.(2021)Lu, Ma, Yang, Zhang, Liu, Chu, Yan, and
  Ouyang]{lu2021geometry}
Yan Lu, Xinzhu Ma, Lei Yang, Tianzhu Zhang, Yating Liu, Qi~Chu, Junjie Yan, and
  Wanli Ouyang.
\newblock Geometry uncertainty projection network for monocular 3d object
  detection.
\newblock In \emph{Proceedings of the IEEE/CVF International Conference on
  Computer Vision (ICCV)}, pages 3111--3121, 2021.

\bibitem[Ma et~al.(2019)Ma, Wang, Li, Zhang, Ouyang, and Fan]{ma2019accurate}
Xinzhu Ma, Zhihui Wang, Haojie Li, Pengbo Zhang, Wanli Ouyang, and Xin Fan.
\newblock Accurate monocular 3d object detection via color-embedded 3d
  reconstruction for autonomous driving.
\newblock In \emph{Proceedings of the IEEE/CVF International Conference on
  Computer Vision (ICCV)}, pages 6851--6860, 2019.

\bibitem[Ma et~al.(2020)Ma, Liu, Xia, Zhang, Zeng, and
  Ouyang]{ma2020rethinking}
Xinzhu Ma, Shinan Liu, Zhiyi Xia, Hongwen Zhang, Xingyu Zeng, and Wanli Ouyang.
\newblock Rethinking pseudo-lidar representation.
\newblock In \emph{European Conference on Computer Vision (ECCV)}, pages
  311--327, 2020.

\bibitem[Mao et~al.(2021)Mao, Xue, Niu, Bai, Feng, Liang, Xu, and
  Xu]{mao2021voxel}
Jiageng Mao, Yujing Xue, Minzhe Niu, Haoyue Bai, Jiashi Feng, Xiaodan Liang,
  Hang Xu, and Chunjing Xu.
\newblock Voxel transformer for 3d object detection.
\newblock In \emph{Proceedings of the IEEE/CVF International Conference on
  Computer Vision}, pages 3164--3173, 2021.

\bibitem[Misra et~al.(2021)Misra, Girdhar, and Joulin]{misra2021end}
Ishan Misra, Rohit Girdhar, and Armand Joulin.
\newblock An end-to-end transformer model for 3d object detection.
\newblock In \emph{Proceedings of the IEEE/CVF International Conference on
  Computer Vision}, pages 2906--2917, 2021.

\bibitem[Pan et~al.(2020)Pan, Xia, Song, Li, and Huang]{pan20203d}
Xuran Pan, Zhuofan Xia, Shiji Song, Li~Erran Li, and Gao Huang.
\newblock 3d object detection with pointformer.
\newblock \emph{arXiv preprint arXiv:2012.11409}, 2020.

\bibitem[Philion and Fidler(2020)]{philion2020lift}
Jonah Philion and Sanja Fidler.
\newblock Lift, splat, shoot: Encoding images from arbitrary camera rigs by
  implicitly unprojecting to 3d.
\newblock In \emph{Computer Vision--ECCV 2020: 16th European Conference,
  Glasgow, UK, August 23--28, 2020, Proceedings, Part XIV 16}, pages 194--210.
  Springer, 2020.

\bibitem[Qi et~al.(2017{\natexlab{a}})Qi, Su, Mo, and Guibas]{qi2017pointnet}
Charles~R Qi, Hao Su, Kaichun Mo, and Leonidas~J Guibas.
\newblock Pointnet: Deep learning on point sets for 3d classification and
  segmentation.
\newblock In \emph{Proceedings of the IEEE conference on computer vision and
  pattern recognition (CVPR)}, pages 652--660, 2017{\natexlab{a}}.

\bibitem[Qi et~al.(2017{\natexlab{b}})Qi, Yi, Su, and Guibas]{qi2017pointnet++}
Charles~Ruizhongtai Qi, Li~Yi, Hao Su, and Leonidas~J Guibas.
\newblock Pointnet++: Deep hierarchical feature learning on point sets in a
  metric space.
\newblock \emph{Advances in Neural Information Processing Systems (NIPS)},
  30:\penalty0 5099--5108, 2017{\natexlab{b}}.

\bibitem[Reading et~al.(2021)Reading, Harakeh, Chae, and
  Waslander]{reading2021categorical}
Cody Reading, Ali Harakeh, Julia Chae, and Steven~L Waslander.
\newblock Categorical depth distribution network for monocular 3d object
  detection.
\newblock In \emph{Proceedings of the IEEE/CVF Conference on Computer Vision
  and Pattern Recognition (CVPR)}, pages 8555--8564, 2021.

\bibitem[Sheng et~al.(2021)Sheng, Cai, Liu, Deng, Huang, Hua, and
  Zhao]{sheng2021improving}
Hualian Sheng, Sijia Cai, Yuan Liu, Bing Deng, Jianqiang Huang, Xian-Sheng Hua,
  and Min-Jian Zhao.
\newblock Improving 3d object detection with channel-wise transformer.
\newblock In \emph{Proceedings of the IEEE/CVF International Conference on
  Computer Vision (ICCV)}, pages 2743--2752, 2021.

\bibitem[Sheng et~al.(2022)Sheng, Cai, Zhao, Deng, Huang, Hua, Zhao, and
  Lee]{sheng2022rethinking}
Hualian Sheng, Sijia Cai, Na~Zhao, Bing Deng, Jianqiang Huang, Xian-Sheng Hua,
  Min-Jian Zhao, and Gim~Hee Lee.
\newblock Rethinking iou-based optimization for single-stage 3d object
  detection.
\newblock In \emph{Proceedings of the European Conference on Computer Vision},
  pages 544--561. Springer, 2022.

\bibitem[Sheng et~al.(2023)Sheng, Cai, Zhao, Deng, Zhao, and Lee]{sheng2023pdr}
Hualian Sheng, Sijia Cai, Na~Zhao, Bing Deng, Min-Jian Zhao, and Gim~Hee Lee.
\newblock Pdr: Progressive depth regularization for monocular 3d object
  detection.
\newblock \emph{IEEE Transactions on Circuits and Systems for Video
  Technology}, 2023.

\bibitem[Shi et~al.(2019)Shi, Wang, and Li]{shi2019pointrcnn}
Shaoshuai Shi, Xiaogang Wang, and Hongsheng Li.
\newblock Pointrcnn: 3d object proposal generation and detection from point
  cloud.
\newblock In \emph{Proceedings of the IEEE Conference on Computer Vision and
  Pattern Recognition (CVPR)}, pages 770--779, 2019.

\bibitem[Shi et~al.(2020{\natexlab{a}})Shi, Guo, Jiang, Wang, Shi, Wang, and
  Li]{shi2020pv}
Shaoshuai Shi, Chaoxu Guo, Li~Jiang, Zhe Wang, Jianping Shi, Xiaogang Wang, and
  Hongsheng Li.
\newblock Pv-rcnn: Point-voxel feature set abstraction for 3d object detection.
\newblock In \emph{Proceedings of the IEEE/CVF Conference on Computer Vision
  and Pattern Recognition}, pages 10529--10538, 2020{\natexlab{a}}.

\bibitem[Shi et~al.(2020{\natexlab{b}})Shi, Wang, Shi, Wang, and
  Li]{shi2020points}
Shaoshuai Shi, Zhe Wang, Jianping Shi, Xiaogang Wang, and Hongsheng Li.
\newblock From points to parts: 3d object detection from point cloud with
  part-aware and part-aggregation network.
\newblock \emph{IEEE Transactions on Pattern Analysis and Machine
  Intelligence}, 43\penalty0 (8):\penalty0 2647--2664, 2020{\natexlab{b}}.

\bibitem[Shi et~al.(2023)Shi, Jiang, Deng, Wang, Guo, Shi, Wang, and
  Li]{shi2023pv}
Shaoshuai Shi, Li~Jiang, Jiajun Deng, Zhe Wang, Chaoxu Guo, Jianping Shi,
  Xiaogang Wang, and Hongsheng Li.
\newblock Pv-rcnn++: Point-voxel feature set abstraction with local vector
  representation for 3d object detection.
\newblock \emph{International Journal of Computer Vision}, 131\penalty0
  (2):\penalty0 531--551, 2023.

\bibitem[Shi et~al.(2021)Shi, Ye, Chen, Chen, Chen, and Kim]{shi2021geometry}
Xuepeng Shi, Qi~Ye, Xiaozhi Chen, Chuangrong Chen, Zhixiang Chen, and Tae-Kyun
  Kim.
\newblock Geometry-based distance decomposition for monocular 3d object
  detection.
\newblock In \emph{Proceedings of the IEEE/CVF International Conference on
  Computer Vision (ICCV)}, pages 15172--15181, 2021.

\bibitem[Song et~al.(2021)Song, Lim, and Kim]{song2021monocular}
Minsoo Song, Seokjae Lim, and Wonjun Kim.
\newblock Monocular depth estimation using laplacian pyramid-based depth
  residuals.
\newblock \emph{IEEE Transactions on Circuits and Systems for Video Technology
  (TCSVT)}, 31\penalty0 (11):\penalty0 4381--4393, 2021.

\bibitem[Song and Xiao(2016)]{song2016deep}
Shuran Song and Jianxiong Xiao.
\newblock Deep sliding shapes for amodal 3d object detection in rgb-d images.
\newblock In \emph{Proceedings of the IEEE Conference on Computer Vision and
  Pattern Recognition (CVPR)}, pages 808--816, 2016.

\bibitem[Team(2020)]{openpcdet2020}
OpenPCDet~Development Team.
\newblock Openpcdet: An open-source toolbox for 3d object detection from point
  clouds.
\newblock \url{https://github.com/open-mmlab/OpenPCDet}, 2020.

\bibitem[Touvron et~al.(2021)Touvron, Cord, Douze, Massa, Sablayrolles, and
  J{\'e}gou]{touvron2021training}
Hugo Touvron, Matthieu Cord, Matthijs Douze, Francisco Massa, Alexandre
  Sablayrolles, and Herv{\'e} J{\'e}gou.
\newblock Training data-efficient image transformers \& distillation through
  attention.
\newblock In \emph{International Conference on Machine Learning}, pages
  10347--10357. PMLR, 2021.

\bibitem[Vaswani et~al.(2017)Vaswani, Shazeer, Parmar, Uszkoreit, Jones, Gomez,
  Kaiser, and Polosukhin]{vaswani2017attention}
Ashish Vaswani, Noam Shazeer, Niki Parmar, Jakob Uszkoreit, Llion Jones,
  Aidan~N Gomez, {\L}ukasz Kaiser, and Illia Polosukhin.
\newblock Attention is all you need.
\newblock In \emph{Advances in Neural Information Processing Systems (NIPS)},
  pages 5998--6008, 2017.

\bibitem[Vyas et~al.(2020)Vyas, Katharopoulos, and Fleuret]{vyas2020fast}
Apoorv Vyas, Angelos Katharopoulos, and Fran{\c{c}}ois Fleuret.
\newblock Fast transformers with clustered attention.
\newblock \emph{Advances in Neural Information Processing Systems},
  33:\penalty0 21665--21674, 2020.

\bibitem[Wang et~al.(2021{\natexlab{a}})Wang, Zhang, Zhu, Zhang, He, Li, and
  Xue]{wang2021progressive}
Li~Wang, Li~Zhang, Yi~Zhu, Zhi Zhang, Tong He, Mu~Li, and Xiangyang Xue.
\newblock Progressive coordinate transforms for monocular 3d object detection.
\newblock \emph{Advances in Neural Information Processing Systems (NeurIPS)},
  34, 2021{\natexlab{a}}.

\bibitem[Wang et~al.(2020)Wang, Li, Khabsa, Fang, and Ma]{wang2020linformer}
Sinong Wang, Belinda~Z Li, Madian Khabsa, Han Fang, and Hao Ma.
\newblock Linformer: Self-attention with linear complexity.
\newblock \emph{arXiv preprint arXiv:2006.04768}, 2020.

\bibitem[Wang et~al.(2021{\natexlab{b}})Wang, Zhu, Pang, and
  Lin]{wang2021fcos3d}
Tai Wang, Xinge Zhu, Jiangmiao Pang, and Dahua Lin.
\newblock Fcos3d: Fully convolutional one-stage monocular 3d object detection.
\newblock In \emph{Proceedings of the IEEE/CVF International Conference on
  Computer Vision (ICCV)}, pages 913--922, 2021{\natexlab{b}}.

\bibitem[Wang et~al.(2019)Wang, Chao, Garg, Hariharan, Campbell, and
  Weinberger]{wang2019pseudo}
Yan Wang, Wei-Lun Chao, Divyansh Garg, Bharath Hariharan, Mark Campbell, and
  Kilian~Q Weinberger.
\newblock Pseudo-lidar from visual depth estimation: Bridging the gap in 3d
  object detection for autonomous driving.
\newblock In \emph{Proceedings of the IEEE/CVF Conference on Computer Vision
  and Pattern Recognition (CVPR)}, pages 8445--8453, 2019.

\bibitem[Wang et~al.(2022)Wang, Guizilini, Zhang, Wang, Zhao, and
  Solomon]{wang2022detr3d}
Yue Wang, Vitor~Campagnolo Guizilini, Tianyuan Zhang, Yilun Wang, Hang Zhao,
  and Justin Solomon.
\newblock Detr3d: 3d object detection from multi-view images via 3d-to-2d
  queries.
\newblock In \emph{Conference on Robot Learning}, pages 180--191. PMLR, 2022.

\bibitem[Xu et~al.(2022)Xu, Zhong, and Neumann]{xu2022behind}
Qiangeng Xu, Yiqi Zhong, and Ulrich Neumann.
\newblock Behind the curtain: Learning occluded shapes for 3d object detection.
\newblock \emph{Proceedings of the AAAI Conference on Artificial Intelligence},
  36\penalty0 (3):\penalty0 2893--2901, 2022.

\bibitem[Yan et~al.(2018)Yan, Mao, and Li]{yan2018second}
Yan Yan, Yuxing Mao, and Bo~Li.
\newblock Second: Sparsely embedded convolutional detection.
\newblock \emph{Sensors}, 18\penalty0 (10):\penalty0 3337, 2018.

\bibitem[Yang et~al.(2018)Yang, Luo, and Urtasun]{yang2018pixor}
Bin Yang, Wenjie Luo, and Raquel Urtasun.
\newblock Pixor: Real-time 3d object detection from point clouds.
\newblock In \emph{Proceedings of the IEEE conference on Computer Vision and
  Pattern Recognition (CVPR)}, pages 7652--7660, 2018.

\bibitem[Yang et~al.(2023)Yang, Chen, Tian, Tao, Zhu, Zhang, Huang, Li, Qiao,
  Lu, et~al.]{yang2023bevformer}
Chenyu Yang, Yuntao Chen, Hao Tian, Chenxin Tao, Xizhou Zhu, Zhaoxiang Zhang,
  Gao Huang, Hongyang Li, Yu~Qiao, Lewei Lu, et~al.
\newblock Bevformer v2: Adapting modern image backbones to bird's-eye-view
  recognition via perspective supervision.
\newblock In \emph{Proceedings of the IEEE/CVF Conference on Computer Vision
  and Pattern Recognition}, pages 17830--17839, 2023.

\bibitem[Yang et~al.(2020)Yang, Sun, Liu, and Jia]{yang20203dssd}
Zetong Yang, Yanan Sun, Shu Liu, and Jiaya Jia.
\newblock 3dssd: Point-based 3d single stage object detector.
\newblock In \emph{Proceedings of the IEEE/CVF Conference on Computer Vision
  and Pattern Recognition}, pages 11040--11048, 2020.

\bibitem[Yin et~al.(2021)Yin, Zhou, and Krahenbuhl]{yin2021center}
Tianwei Yin, Xingyi Zhou, and Philipp Krahenbuhl.
\newblock Center-based 3d object detection and tracking.
\newblock In \emph{Proceedings of the IEEE/CVF Conference on Computer Vision
  and Pattern Recognition (CVPR)}, pages 11784--11793, 2021.

\bibitem[Zhang et~al.(2023)Zhang, Qiu, Wang, Guo, Cui, Qiao, Li, and
  Gao]{zhang2023monodetr}
Renrui Zhang, Han Qiu, Tai Wang, Ziyu Guo, Ziteng Cui, Yu~Qiao, Hongsheng Li,
  and Peng Gao.
\newblock Monodetr: Depth-guided transformer for monocular 3d object detection.
\newblock In \emph{Proceedings of the IEEE/CVF International Conference on
  Computer Vision}, pages 9155--9166, 2023.

\bibitem[Zhang et~al.(2022)Zhang, Hu, Xu, Ma, Wan, and Guo]{zhang2022notall}
Yifan Zhang, Qingyong Hu, Guoquan Xu, Yanxin Ma, Jianwei Wan, and Yulan Guo.
\newblock Not all points are equal: Learning highly efficient point-based
  detectors for 3d lidar point clouds.
\newblock In \emph{Proceedings of the IEEE/CVF Conference on Computer Vision
  and Pattern Recognition}, pages 18953--18962, 2022.

\bibitem[Zhang et~al.(2021)Zhang, Lu, and Zhou]{zhang2021objects}
Yunpeng Zhang, Jiwen Lu, and Jie Zhou.
\newblock Objects are different: Flexible monocular 3d object detection.
\newblock In \emph{Proceedings of the IEEE/CVF Conference on Computer Vision
  and Pattern Recognition (CVPR)}, pages 3289--3298, 2021.

\bibitem[Zhao et~al.(2021)Zhao, Jiang, Jia, Torr, and Koltun]{zhao2021point}
Hengshuang Zhao, Li~Jiang, Jiaya Jia, Philip~HS Torr, and Vladlen Koltun.
\newblock Point transformer.
\newblock In \emph{Proceedings of the IEEE/CVF International Conference on
  Computer Vision}, pages 16259--16268, 2021.

\bibitem[Zheng et~al.(2021)Zheng, Tang, Chen, Jiang, and Fu]{zheng2021cia}
Wu~Zheng, Weiliang Tang, Sijin Chen, Li~Jiang, and Chi-Wing Fu.
\newblock Cia-ssd: Confident iou-aware single-stage object detector from point
  cloud.
\newblock \emph{Proceedings of the AAAI Conference on Artificial Intelligence},
  35\penalty0 (4):\penalty0 3555--3562, 2021.

\bibitem[Zhou et~al.(2019)Zhou, Zhuo, and Krahenbuhl]{zhou2019bottom}
Xingyi Zhou, Jiacheng Zhuo, and Philipp Krahenbuhl.
\newblock Bottom-up object detection by grouping extreme and center points.
\newblock In \emph{Proceedings of the IEEE Conference on Computer Vision and
  Pattern Recognition (CVPR)}, pages 850--859, 2019.

\bibitem[Zhou and Tuzel(2018)]{zhou2018voxelnet}
Yin Zhou and Oncel Tuzel.
\newblock Voxelnet: End-to-end learning for point cloud based 3d object
  detection.
\newblock In \emph{Proceedings of the IEEE/CVF Conference on Computer Vision
  and Pattern Recognition}, pages 4490--4499, 2018.

\end{thebibliography}

\newpage

\end{document}